\documentclass[11pt]{article}

\usepackage[final]{acl}

\usepackage[T1]{fontenc}
\usepackage[utf8]{inputenc}
\usepackage{newtxtext,newtxmath}  
\usepackage{microtype}
\usepackage{inconsolata}
\usepackage{graphicx}
\usepackage{booktabs}
\usepackage{tabularx}
\usepackage{multirow}
\usepackage{amsmath}
\usepackage{amssymb}
\usepackage{stfloats}
\usepackage{float}
\usepackage{subcaption}
\usepackage[framemethod=default]{mdframed}
\usepackage{tikz}
\usetikzlibrary{positioning,arrows.meta,fit,backgrounds,shapes.geometric,calc}
\definecolor{clrGemcoA}{HTML}{B0CA97}   
\definecolor{clrGemcoB}{HTML}{88C4C8}   
\definecolor{clrReal}{HTML}{D1BC8A}   
\definecolor{clrResult}{HTML}{E8A087} 
\definecolor{clrEmotion}{HTML}{CB7FA6} 
\definecolor{clrSubst}{HTML}{9F92CC}   
\usepackage[most]{tcolorbox}
\usepackage{enumitem}
\definecolor{clrPromptDE}{HTML}{2A9D8F} 
\definecolor{clrPromptEN}{HTML}{3D5A80} 

\IfFileExists{oncoco_macros/gold_macros.tex}{%

}{}
\IfFileExists{oncoco_macros/lexical_macros.tex}{%

\newcommand{\jaccSplit}{.350}
\newcommand{\jaccSplitSD}{.003}
\newcommand{\jaccRPCo}{.342}
\newcommand{\jaccSplitCo}{.346}
\newcommand{\jaccSplitCoSD}{.004}
\newcommand{\jaccRPCl}{.302}
\newcommand{\jaccSplitCl}{.322}
\newcommand{\jaccSplitClSD}{.004}
}{%
  \newcommand{\jaccSplit}{.350}\newcommand{\jaccSplitSD}{.003}%
  \newcommand{\jaccRPCo}{.342}\newcommand{\jaccSplitCo}{.346}\newcommand{\jaccSplitCoSD}{.004}%
  \newcommand{\jaccRPCl}{.302}\newcommand{\jaccSplitCl}{.322}\newcommand{\jaccSplitClSD}{.004}%
}
\IfFileExists{oncoco_macros/reanalysis_macros.tex}{%
\newcommand{\NmcoMean}{.0009}

\newcommand{\NmcoPNF}{.0020}
\newcommand{\NmEkMean}{.0008}

\newcommand{\NmEkPNF}{.0017}
\newcommand{\NmLalkMean}{.0034}

\newcommand{\NmLalkPNF}{.0057}
\newcommand{\NmAllMean}{.0039}

\newcommand{\NmAllPNF}{.0065}

\newcommand{\fdrQ}{0.10}

}{%
  \newcommand{\NmcoMean}{.0009}%
  \newcommand{\NmcoPNF}{.0020}\newcommand{\NmEkMean}{.0008}%
  \newcommand{\NmEkPNF}{.0017}%
  \newcommand{\NmLalkMean}{.0034}%
  \newcommand{\NmLalkPNF}{.0057}\newcommand{\NmAllMean}{.0039}%
  \newcommand{\NmAllPNF}{.0065}%
  \newcommand{\fdrQ}{0.10}%
}
\IfFileExists{oncoco_macros/mauve_macros.tex}{%
\newcommand{\mauveRP}{.560}
\newcommand{\mauveA}{.654}
\newcommand{\mauveB}{.195}
\newcommand{\mauveRR}{.940}
\newcommand{\mauveRRsd}{.030}
\newcommand{\mauveRRmin}{.846}
\newcommand{\mauvePEsconv}{.004}
\newcommand{\mauveREsconv}{.005}
\newcommand{\mauvePAnnomi}{.004}
\newcommand{\mauveRAnnomi}{.004}
\newcommand{\mauveBLenmatch}{.196}
\newcommand{\mauveRealShortLong}{.060}
}{%
  \newcommand{\mauveRP}{.560}\newcommand{\mauveA}{.654}%
  \newcommand{\mauveB}{.195}\newcommand{\mauveRR}{.940}%
  \newcommand{\mauveRRsd}{.030}\newcommand{\mauveRRmin}{.846}%
  \newcommand{\mauvePEsconv}{.004}\newcommand{\mauveREsconv}{.005}%
  \newcommand{\mauvePAnnomi}{.004}\newcommand{\mauveRAnnomi}{.004}%
  \newcommand{\mauveBLenmatch}{.196}\newcommand{\mauveRealShortLong}{.060}%
}
\IfFileExists{oncoco_macros/mauve_genval_macros.tex}{%
\newcommand{\mauveHumanSft}{.907}
\newcommand{\mauveHumanInstruct}{.254}
\newcommand{\mauveSftInstruct}{.511}
\newcommand{\mauveHH}{.818}
\newcommand{\mauveHHsd}{.074}
\newcommand{\mauveXSft}{.742}
\newcommand{\mauveXSftSd}{.092}
\newcommand{\mauveXInstruct}{.214}
\newcommand{\mauveXInstructSd}{.056}
}{%
  \newcommand{\mauveHumanSft}{.907}\newcommand{\mauveHumanInstruct}{.254}%
  \newcommand{\mauveSftInstruct}{.511}%
  \newcommand{\mauveHH}{.818}\newcommand{\mauveHHsd}{.074}%
  \newcommand{\mauveXSft}{.742}\newcommand{\mauveXSftSd}{.092}%
  \newcommand{\mauveXInstruct}{.214}\newcommand{\mauveXInstructSd}{.056}%
}
\IfFileExists{oncoco_macros/dyad_macros.tex}{%

\newcommand{\lsmProxyMed}{.768}
\newcommand{\lsmRealMed}{.794}

\newcommand{\lsmKS}{.312}
\newcommand{\lsmNullPNF}{.18}
}{%
  \newcommand{\lsmProxyMed}{.768}\newcommand{\lsmRealMed}{.794}%
  \newcommand{\lsmKS}{.312}\newcommand{\lsmNullPNF}{.18}%
}
\IfFileExists{oncoco_macros/splithalf_macros.tex}{%
\newcommand{\shRPCo}{.0035}
\newcommand{\shRRCo}{.0016}
\newcommand{\shPctlCo}{96}
\newcommand{\shRPCoLeaf}{.0086}
\newcommand{\shRRCoLeaf}{.0051}
\newcommand{\shPctlCoLeaf}{96}
\newcommand{\shRPAll}{.0159}
\newcommand{\shRRAll}{.0059}
\newcommand{\shPctlAll}{$>$99}
\newcommand{\shRPEk}{.0036}
\newcommand{\shRREk}{.0013}
\newcommand{\shPctlEk}{99}
\newcommand{\shRPGo}{.0087}
\newcommand{\shRRGo}{.0057}
\newcommand{\shPctlGo}{95}
}{%
  \newcommand{\shRPCo}{.0035}\newcommand{\shRRCo}{.0016}\newcommand{\shPctlCo}{96}%
  \newcommand{\shRPCoLeaf}{.0086}\newcommand{\shRRCoLeaf}{.0051}\newcommand{\shPctlCoLeaf}{96}%
  \newcommand{\shRPAll}{.0159}\newcommand{\shRRAll}{.0059}\newcommand{\shPctlAll}{$>$99}%
  \newcommand{\shRPEk}{.0036}\newcommand{\shRREk}{.0013}\newcommand{\shPctlEk}{99}%
  \newcommand{\shRPGo}{.0087}\newcommand{\shRRGo}{.0057}\newcommand{\shPctlGo}{95}%
}
\IfFileExists{oncoco_macros/table_stats_macros.tex}{%
\newcommand{\ciCoLo}{.0022}
\newcommand{\ciCoHi}{.0070}
\newcommand{\vCo}{.07}

\newcommand{\ciEkLo}{.0016}
\newcommand{\ciEkHi}{.0082}
\newcommand{\vEk}{.07}

\newcommand{\vGo}{.11}

\newcommand{\perbinRowsApp}{%
0--20\,\% & .008 & .11 & .024 & .18 & .008 & .11 & .030 & .20 \\
20--40\,\% & .009 & .11 & .021 & .17 & .005 & .08 & .020 & .17 \\
40--60\,\% & .002 & .06 & .020 & .16 & .004 & .08 & .015 & .14 \\
60--80\,\% & .011 & .12 & .025 & .18 & .011 & .12 & .022 & .17 \\
80--100\,\% & .021 & .17 & .032 & .21 & .002 & .06 & .020 & .16 \\
\midrule
Corpus & .004 & .07 & .009 & .11 & .004 & .07 & .009 & .11 \\}
\newcommand{\perbinSplitRows}{%
All (corpus) & .0046 & \textbf{.0066} & .0037 & .0019 & \textbf{.0088} & .0026 \\
0--20\,\% & .0044 & .0117 & .0030 & .0056 & \textbf{.0305} & .0094 \\
20--40\,\% & .0053 & .0082 & .0041 & .0021 & \textbf{.0194} & .0043 \\
40--60\,\% & .0075 & .0067 & .0048 & .0038 & .0012 & .0020 \\
60--80\,\% & .0098 & \textbf{.0348} & .0141 & .0056 & .0096 & .0061 \\
80--100\,\% & .0061 & \textbf{.0391} & .0101 & .0040 & .0104 & .0024 \\}
}{%
  \newcommand{\ciCoLo}{.002}\newcommand{\ciCoHi}{.007}%
  \newcommand{\vCo}{.07}%
  \newcommand{\ciEkLo}{.002}\newcommand{\ciEkHi}{.008}%
  \newcommand{\vEk}{.07}%
  \newcommand{\vGo}{.11}%
  \newcommand{\perbinSplitRows}{%
All (corpus) & .0046 & \textbf{.0066} & .0037 & .0019 & \textbf{.0088} & .0026 \\
0--20\,\% & .0044 & .0117 & .0030 & .0056 & \textbf{.0305} & .0094 \\
20--40\,\% & .0053 & .0082 & .0041 & .0021 & \textbf{.0194} & .0043 \\
40--60\,\% & .0075 & .0067 & .0048 & .0038 & .0012 & .0020 \\
60--80\,\% & .0098 & \textbf{.0348} & .0141 & .0056 & .0096 & .0061 \\
80--100\,\% & .0061 & \textbf{.0391} & .0101 & .0040 & .0104 & .0024 \\}%
  \newcommand{\perbinRowsApp}{%
0--20\,\% & .008 & .11 & .024 & .18 & .008 & .11 & .030 & .20 \\
20--40\,\% & .009 & .11 & .021 & .17 & .005 & .08 & .020 & .17 \\
40--60\,\% & .002 & .06 & .020 & .16 & .004 & .08 & .015 & .14 \\
60--80\,\% & .011 & .12 & .025 & .18 & .011 & .12 & .022 & .17 \\
80--100\,\% & .021 & .17 & .032 & .21 & .002 & .06 & .020 & .16 \\
\midrule
Corpus & .004 & .07 & .009 & .11 & .004 & .07 & .009 & .11 \\}%
}
\IfFileExists{oncoco_macros/downstream_macros.tex}{%
\newcommand{\dsNturns}{325}
\newcommand{\dsNreal}{124}
\newcommand{\dsNjudg}{1{,}950}
\newcommand{\dsHumanWRlo}{68}
\newcommand{\dsHumanWRhi}{77}
\newcommand{\dsProxyWRlo}{41}
\newcommand{\dsProxyWRhi}{48}
\newcommand{\dsInstructWRlo}{29}
\newcommand{\dsInstructWRhi}{37}
\newcommand{\dsHumanWR}{73}
\newcommand{\dsProxyWR}{44}
\newcommand{\dsInstructWR}{33}
\newcommand{\dsProxyBeatsInstruct}{57}
\newcommand{\dsHumanBeatsProxy}{68}
\newcommand{\dsHumanBeatsInstruct}{77}
\newcommand{\dsConsistency}{87}

\newcommand{\dsBThuman}{1.71}
\newcommand{\dsBTproxy}{0.75}
\newcommand{\dsBTinstruct}{0.54}
}{%
  \newcommand{\dsNturns}{325}\newcommand{\dsNreal}{124}%
  \newcommand{\dsHumanWR}{73}\newcommand{\dsProxyWR}{44}\newcommand{\dsInstructWR}{33}%
  \newcommand{\dsProxyBeatsInstruct}{57}\newcommand{\dsHumanBeatsProxy}{68}%
  \newcommand{\dsHumanBeatsInstruct}{77}\newcommand{\dsConsistency}{87}%
  \newcommand{\dsBThuman}{1.71}\newcommand{\dsBTproxy}{0.75}\newcommand{\dsBTinstruct}{0.54}%
  \newcommand{\dsNjudg}{1{,}950}%
  \newcommand{\dsHumanWRlo}{68}\newcommand{\dsHumanWRhi}{77}%
  \newcommand{\dsProxyWRlo}{41}\newcommand{\dsProxyWRhi}{48}%
  \newcommand{\dsInstructWRlo}{29}\newcommand{\dsInstructWRhi}{37}%
}
\IfFileExists{oncoco_macros/transition_macros.tex}{%

\newcommand{\tauMedCoA}{3.8}
\newcommand{\tauMedCoB}{5.8}
}{%
  \newcommand{\tauMedCoA}{3.8}\newcommand{\tauMedCoB}{5.8}%
}
\IfFileExists{oncoco_macros/emotion_transition_macros.tex}{%
\newcommand{\nFdrEkA}{1}

\newcommand{\nFdrEkB}{6}

}{%
  \newcommand{\nFdrEkA}{1}\newcommand{\nFdrEkB}{6}%
}

\title{GEMCo: A Validated, Ethically Releasable Proxy for Inaccessible Counselling Data}

\author{\normalfont
  Philipp Steigerwald$^1$ \quad Eric Rudolph$^1$ \quad Mara Stieler$^2$ \\
  Jennifer Burghardt$^2$ \quad Robert Lehmann$^2$ \quad Jens Albrecht$^1$ \\
  $^1$Faculty of Computer Science, Center for Artificial Intelligence (KIZ) \\
  $^2$Faculty of Social Sciences, Institute for E-Counselling \\
  Technische Hochschule N\"urnberg Georg Simon Ohm, N\"urnberg, Germany \\
  \texttt{\{philipp.steigerwald, eric.rudolph, mara.stieler,}\\
  \texttt{jennifer.burghardt, robert.lehmann, jens.albrecht\}@th-nuernberg.de}%
}

\begin{document}
\maketitle

\begin{abstract}
This paper presents \textbf{GEMCo}, a releasable, human-written proxy for inaccessible counselling data: 86 complete German e-mail counselling conversations (728 messages), expert-authored cases and counsellor sessions with trained role-players.\footnote{Data: \href{https://github.com/th-nuernberg/GEMCo}{github.com/th-nuernberg/GEMCo}.} It is validated against a held-out reference of 124 real counselling conversations.
The proxy and the real conversations are measured against each other on counsellor strategies and client emotions.
The gap is detectable but small.
A generative validation supports the analysis.
The corpus is the primary contribution.
The validation method generalises to any domain where real data cannot be shared but a human-made proxy can.
Privacy and ethics keep real counselling data closed.
GEMCo carries none by design and can be released, a first step toward language research in this domain.
\end{abstract}

\section{Introduction}
\label{sec:intro}

Text-based counselling plays an increasingly important role in mental health service delivery \cite{engelhardt2021}, yet computational tools for counsellor training, quality assurance or assistive systems like intake and triage bots require data that is rarely available due to its sensitive nature \cite{malgaroli2023nlp}.

Almost all available dialogue resources are English (Table~\ref{tab:comparison}), running from crisis conversations to transcribed professional sessions.
Crisis Text Line holds text-message support at scale under restricted access \cite{althoff2016counseling}, while ESConv openly releases crowdsourced emotional-support chats \cite{liu2021ESConv}.
Closer to professional practice is DAIC-WOZ, which records clinical interviews \cite{gratch2014daic}.
AnnoMI and HOPE are transcripts of counselling demonstrations by professionals \cite{wu2022annomi,malhotra2022hope}.
PsyQA is large but single-turn, pairing each question with a long counselling answer rather than a multi-turn dialogue \cite{sun2021psyqa}.

German offers far less.
SMHD-GER collects social-media posts rather than counselling dialogue \cite{zanwar2023smhdger}.
OnCoCo~1.0 annotates single online-counselling messages rather than full threads \cite{oncoco2026}.
German is spoken by over 100 million people, yet no natively German corpus of asynchronous multi-turn counselling exists.
Ethical obligations toward real clients keep authentic counselling data out of public release.

\begin{table}[t]
\centering
\resizebox{\columnwidth}{!}{%
\footnotesize
\begin{tabular}{@{}lcrccl@{}}
\toprule
\textbf{Dataset} & \textbf{Lang.} & \textbf{N} & \textbf{Mt.} & \textbf{Pr.} & \textbf{Public} \\
\midrule
Crisis Text Line \cite{althoff2016counseling} & EN & 80K+ & \checkmark & -- & restricted \\
ESConv \cite{liu2021ESConv} & EN & 1{,}053 & \checkmark & -- & \checkmark \\
DAIC-WOZ \cite{gratch2014daic} & EN & 189 & \checkmark & \checkmark & restricted \\
AnnoMI \cite{wu2022annomi} & EN & 133 & \checkmark & \checkmark & \checkmark \\
HOPE \cite{malhotra2022hope} & EN & 212 & \checkmark & \checkmark & on req.
\\
PsyQA \cite{sun2021psyqa} & ZH & 22K & -- & -- & on req.
\\
SMHD-GER \cite{zanwar2023smhdger} & DE & -- & -- & -- & restricted \\
OnCoCo \cite{oncoco2026} & DE/EN & 2{,}778\textsuperscript{*} & -- & \checkmark & \checkmark \\
\midrule
\textbf{GEMCo} & \textbf{DE} & \textbf{86} & \checkmark & \checkmark & \checkmark \\
\bottomrule
\end{tabular}%
}
\caption{Mental-health NLP dialogue resources.
Mt.\ = multi-turn; Pr.\ = professionals; \checkmark{} = yes, -- = no.
\textsuperscript{*}Single messages, not full threads.}
\label{tab:comparison}
\end{table}

This paper makes three contributions.
\begin{enumerate}[label=(\roman*),leftmargin=2.1em,itemsep=1.5pt,topsep=2pt,parsep=0pt]
\item \textbf{GEMCo}, the first publicly available German corpus of asynchronous multi-turn e-mail counselling for mental-health support: 86 complete, human-written threads (GEMCo-A, expert-authored cases; GEMCo-B, counsellor sessions with trained role-players), released under CC~BY~4.0 as a \textit{proxy} for inaccessible counselling data.
\item A generalisable method for validating such a proxy. A releasable, human-made proxy is compared with real data that cannot be shared. The gap is scaled by the real data’s own split-half resampling noise, so it reads as inside or beyond natural variation. It needs only aggregate label distributions of the withheld set, so it transfers to any sensitive domain.
\item The validation itself, run on counselling. GEMCo-A, GEMCo-B and their pool are compared with the 124 withheld real conversations at corpus, progress and message level, with a generative probe of the training signal.
\end{enumerate}

The corpus is the main contribution.
The method makes its release defensible and transfers beyond counselling.

\section{Corpus Description}
\label{sec:corpus}

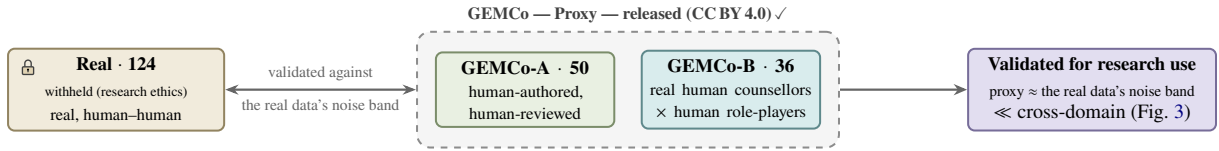
\begin{figure*}[t]
\centering
\resizebox{\textwidth}{!}{%
\begin{tikzpicture}[
  font=\footnotesize,
  every node/.style={align=center},
  modbox/.style={rectangle, rounded corners=3pt, line width=0.9pt,
                 minimum width=2.3cm, minimum height=0.98cm},
  pipe/.style={-{Stealth[length=6pt,width=5pt]}, line width=1.0pt, black!60},
  bidir/.style={{Stealth[length=6pt]}-{Stealth[length=6pt]}, line width=1.0pt, black!60},
  arlbl/.style={font=\fontsize{7}{8.5}\selectfont, text=black!60},
]
\node[modbox, draw=clrReal!70!black, fill=clrReal!30, minimum width=3.4cm] (real)
   {\fontsize{8.4}{9.6}\selectfont\textbf{Real $\cdot$ 124}\\[2pt]\fontsize{6.2}{7.2}\selectfont withheld (research ethics)\\[1pt]\fontsize{7.4}{8.6}\selectfont real, human--human};
\begin{scope}[shift={($(real.north west)+(0.33,-0.30)$)}, draw=black!75, line width=0.6pt]
  \draw (-0.05,0.0) -- (-0.05,0.055) arc(180:0:0.05) -- (0.05,0.0);   
  \draw[fill=clrReal!65, rounded corners=0.5pt] (-0.088,-0.12) rectangle (0.088,0.0); 
  \fill (0,-0.045) circle (0.017);                                    
  \draw (0,-0.05) -- (0,-0.085);
\end{scope}
\node[modbox, text width=2.55cm, draw=clrGemcoA!70!black, fill=clrGemcoA!28,
      right=3.3cm of real] (suba)
   {\fontsize{8.4}{9.6}\selectfont\textbf{GEMCo-A $\cdot$ 50}\\[2pt]\fontsize{7.4}{8.6}\selectfont human-authored,\\ human-reviewed};
\node[modbox, text width=2.55cm, draw=clrGemcoB!70!black, fill=clrGemcoB!32, right=4mm of suba] (subb)
   {\fontsize{8.4}{9.6}\selectfont\textbf{GEMCo-B $\cdot$ 36}\\[2pt]\fontsize{7.4}{8.6}\selectfont real human counsellors\\ $\times$ human role-players};
\begin{scope}[on background layer]
\node[draw=black!45, dashed, rounded corners=6pt, line width=0.9pt,
      fill=black!4, fit=(suba)(subb), inner sep=8pt,
      label={[font=\fontsize{7.6}{9}\selectfont\bfseries, text=black!75]above:GEMCo --- Proxy --- released (CC\,BY\,4.0)\,\checkmark}] (proxy) {};
\end{scope}
\node[modbox, draw=clrSubst!75!black, fill=clrSubst!30, minimum width=3.9cm,
      right=2.0cm of proxy] (out)
   {\fontsize{8.4}{9.6}\selectfont\textbf{Validated for research use}\\[2pt]\fontsize{6.8}{8}\selectfont proxy $\approx$ the real data’s noise band\\ $\ll$ cross-domain (Fig.~\ref{fig:splithalf})};
\draw[bidir] (real.east) -- node[arlbl, above]{validated against} node[arlbl, below]{the real data’s noise band} (proxy.west);
\draw[pipe] (proxy.east) -- (out.west);
\end{tikzpicture}}
\caption{Research design.
The releasable human-generated \textit{proxy} corpus (GEMCo-A\,+\,GEMCo-B) is validated against the withheld \textit{real} reference.}
\label{fig:overview}
\end{figure*}

GEMCo was built at Technische Hochschule N\"urnberg Georg Simon Ohm with active practitioners of German online counselling.
The releasable human-generated proxy corpus is built from two sources.
The expert-authored cases of GEMCo-A come from one of Germany's largest counselling services.
GEMCo-B was created through role-plays involving counsellors from this and two further German counselling services.
Every thread is a complete, human-written asynchronous e-mail exchange and the counsellors were instructed to counsel just as they would in real practice.
A third set of 124 authentic conversations donated by consenting and informed real clients, the held-out \textit{real data}, serves for aggregate validation but cannot be released (Figure~\ref{fig:overview}).
The distinction between \textit{proxy} and \textit{real data} is releasability, not authenticity.
Every message in all three corpora is human-written, none of it machine-generated.
GEMCo-A and GEMCo-B are released in full under CC~BY~4.0.

\subsection{GEMCo-A: Expert Case Authoring}
Two practising counsellors collaboratively authored 50 threads (403 messages), drawing on their experience in online counselling.
Both are graduate social pedagogues with systemic and client-centred training, decades of counselling experience and more than ten years of it in online counselling.
They modelled every case on how a counselling conversation typically begins and on the kinds of first request they meet in practice, then built it up from there.
Both sides were written together and revised iteratively as each case took shape.
Each thread was then read independently for realism by a third professional, a certified online counsellor who is also a social-science researcher.
The cases deliberately span concern types (from parenting, school and relationships to identity, pregnancy, grief, self-harm and family conflict) and communication styles.
GEMCo-A is fictional and holds no real personal data, so it can be released without restriction.

\subsection{GEMCo-B: Role-Playing Sessions}
GEMCo-B was collected during a four-week field deployment~\cite{steigerwald2025caia}
that paired 34~professional counsellors with 13~trained students playing counsellee roles.
Nine counsellors worked in educational and family counselling, fourteen in addiction and eleven had just completed certification.
Each student corresponded with two or three counsellors at once, freely interpreting shared case vignettes (family, youth, addiction), so threads emerged from genuine asynchronous exchanges over weeks.
Several pairs sharing one vignette give controlled thematic overlap.
The resulting 36 threads comprise 325 messages with the stylistic breadth of 34~practitioners, two of whom completed two threads each.
Only the help-seeker role was played, by trained humans rather than a model, so the counselling itself is genuine and the counsellor side fully authentic.
LLM-simulated clients reproduce the communication patterns of counselling role-plays only in part~\cite{rudolph2025roleplay}.
All participants consented to donate their data and every client is role-played, so GEMCo-B carries no confidentiality constraints.

\subsection{The Held-Out Real Data}
The 124 real conversations were obtained through voluntary, revocable data donation with informed consent at a real counselling institution.
The data was anonymised and pseudonymised in-house at the counselling service before any of it reached the research team.
In this paper only aggregate statistics are reported from this actual counselling data.
Structural differences from the proxy are expected (Table~\ref{tab:stats}).
The early drop-offs are the clearest.
Nearly a quarter of the real threads end after at most two messages, typically a client enquiry and the counsellor's reply with no further message.
The other 95 run longer, while the proxy holds 86 complete threads end to end.
Real clients write markedly longer messages (Table~\ref{tab:stats}).
This is a first trace of the client-side contrast the validation later locates.
The real data is never released and serves only as the reference GEMCo is measured against in the sections that follow.
A close match licenses the releasable proxy as an ethically clean stand-in where the real conversations cannot be shared.
Every check uses all 124 real conversations, except where the analysis depends on conversation position or order.
The progress bins (Section~\ref{sec:bins}) and within-message transitions (Section~\ref{sec:dyad}) use the 48 complete threads, where position is well defined.
Filtering is stated and motivated wherever it applies.

\begin{table}[!t]
\centering
\resizebox{\columnwidth}{!}{%
\footnotesize
\begin{tabular}{@{}lrrr@{}}
\toprule
& \textbf{GEMCo-A} & \textbf{GEMCo-B} & \textbf{Real} \\
\midrule
Conversations & 50 & 36 & 124 \\
Messages & 403 & 325 & 587 \\
Words & 118{,}514 & 64{,}660 & 206{,}082 \\
\midrule
Msgs/conv.\ (mean$\pm$sd) & 8.1$\pm$3.8 & 9.0$\pm$4.2 & 4.7$\pm$3.3 \\
\midrule
Words/msg (client) & 214$\pm$163 & 139$\pm$103 & 345$\pm$312 \\
Words/msg (couns.) & 377$\pm$235 & 274$\pm$204 & 356$\pm$331 \\
\midrule
Client msgs & 206\,(51\%) & 180\,(55\%) & 262\,(45\%) \\
Counsellor msgs & 197\,(49\%) & 145\,(45\%) & 325\,(55\%) \\
\bottomrule
\end{tabular}%
}
\caption{Corpus statistics for the two released subcorpora and the withheld real reference.}
\label{tab:stats}
\end{table}

\section{Annotation Pipeline}
\label{sec:annotation}

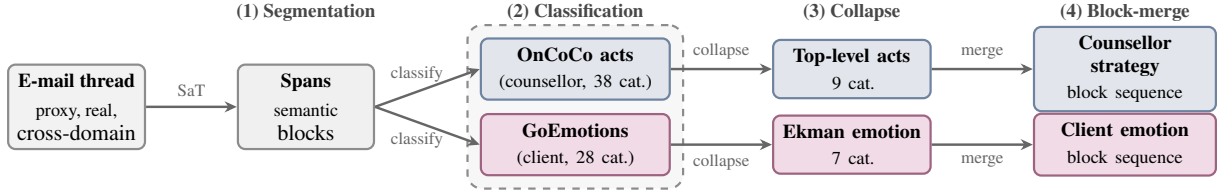
\begin{figure*}[t]
\centering
\resizebox{\textwidth}{!}{%
\begin{tikzpicture}[
  font=\footnotesize,
  every node/.style={align=center},
  modbox/.style={rectangle, rounded corners=3pt, line width=0.9pt,
                 minimum width=2.0cm, minimum height=0.74cm},
  pipe/.style={-{Stealth[length=5pt,width=4.5pt]}, line width=1.0pt, black!60},
  arlbl/.style={font=\fontsize{7}{8.5}\selectfont, text=black!60},
  stage/.style={font=\fontsize{7.8}{9}\selectfont\bfseries, text=black!72},
]
\node[modbox, draw=black!45, fill=black!5, minimum width=1.95cm] (thread)
   {\fontsize{8}{9.4}\selectfont\textbf{E-mail thread}\\[2pt]\fontsize{7.4}{8.6}\selectfont proxy, real,\\ cross-domain};
\node[modbox, draw=black!45, fill=black!6, minimum width=1.95cm, right=1.3cm of thread] (spans)
   {\fontsize{8}{9.4}\selectfont\textbf{Spans}\\[2pt]\fontsize{7.4}{8.6}\selectfont semantic\\ blocks};
\node[modbox, draw=clrPromptEN!75, fill=clrPromptEN!15, text width=2.45cm,
      right=1.5cm of spans, yshift=0.54cm] (oncoco)
   {\fontsize{8}{9.4}\selectfont\textbf{OnCoCo acts}\\[1pt]\fontsize{7.4}{8.6}\selectfont(counsellor, 38 cat.)};
\node[modbox, draw=clrEmotion!78!black, fill=clrEmotion!28, text width=2.45cm,
      right=1.5cm of spans, yshift=-0.54cm] (emot)
   {\fontsize{8}{9.4}\selectfont\textbf{GoEmotions}\\[1pt]\fontsize{7.4}{8.6}\selectfont(client, 28 cat.)};
\node[modbox, draw=clrPromptEN!75, fill=clrPromptEN!15, text width=2.0cm,
      right=1.45cm of oncoco] (colco)
   {\fontsize{8}{9.4}\selectfont\textbf{Top-level acts}\\[1pt]\fontsize{7.4}{8.6}\selectfont 9 cat.};
\node[modbox, draw=clrEmotion!78!black, fill=clrEmotion!28, text width=2.0cm,
      right=1.45cm of emot] (colcl)
   {\fontsize{8}{9.4}\selectfont\textbf{Ekman emotion}\\[1pt]\fontsize{7.4}{8.6}\selectfont 7 cat.};
\node[modbox, draw=clrPromptEN!75, fill=clrPromptEN!15, text width=2.35cm,
      right=1.45cm of colco] (stratblk)
   {\fontsize{8}{9.4}\selectfont\textbf{Counsellor strategy}\\[1pt]\fontsize{7.4}{8.6}\selectfont block sequence};
\node[modbox, draw=clrEmotion!78!black, fill=clrEmotion!28, text width=2.35cm,
      right=1.45cm of colcl] (emoblk)
   {\fontsize{8}{9.4}\selectfont\textbf{Client emotion}\\[1pt]\fontsize{7.4}{8.6}\selectfont block sequence};
\begin{scope}[on background layer]
\node[draw=black!45, dashed, rounded corners=6pt, line width=0.9pt, fill=black!3,
      fit=(oncoco)(emot), inner sep=5pt] (classbox) {};
\end{scope}
\draw[pipe] (thread) -- node[arlbl,above]{SaT} (spans);
\draw[pipe] (spans.east) -- node[arlbl,above,pos=0.4]{classify} (oncoco.west);
\draw[pipe] (spans.east) -- node[arlbl,below,pos=0.4]{classify} (emot.west);
\draw[pipe] (oncoco) -- node[arlbl,above]{collapse} (colco);
\draw[pipe] (emot) -- node[arlbl,below]{collapse} (colcl);
\draw[pipe] (colco) -- node[arlbl,above]{merge} (stratblk);
\draw[pipe] (colcl) -- node[arlbl,below]{merge} (emoblk);
\node[stage] (st2) at ($(classbox.north)+(0,0.17)$) {(2) Classification};
\node[stage] at (st2 -| spans) {(1) Segmentation};
\node[stage] at (st2 -| colco) {(3) Collapse};
\node[stage] at (st2 -| stratblk) {(4) Block-merge};
\end{tikzpicture}}
\caption{Annotation pipeline. Each e-mail is split into spans, classified at the finest OnCoCo (counsellor) and GoEmotions (client) level, collapsed to nine acts / seven emotions and block-merged (\S\ref{sec:annot_seg}--\S\ref{sec:annot_blockmerge}).}
\label{fig:pipeline}
\end{figure*}

Both datasets, the proxy and the real data, pass through one shared pipeline.
It labels spans (semantic blocks) along two dimensions: how counsellors structure their interventions and how client emotions are expressed.
Two English counselling corpora serve later as a cross-domain contrast (Section~\ref{sec:methodology}), the crowdsourced emotional-support chats of ESConv \cite{liu2021ESConv} and the motivational-interviewing transcripts of AnnoMI \cite{wu2022annomi}.
Both were chosen for their proximity to the GEMCo setting, each a mental-health support corpus though neither is asynchronous e-mail counselling.
They pass through the same classifiers, applied to their native turn segmentation (Appendix~\ref{app:baselines}).
The pipeline runs in four stages (Figure~\ref{fig:pipeline}): (1)~segmentation into semantic spans (\S\ref{sec:annot_seg}), (2)~classification of every span at the finest level and (3)~a collapse to the nine-act and seven-emotion analysis level (\S\ref{sec:annot_oncoco},~\S\ref{sec:annot_emotion}), then (4)~a per-taxonomy block-merge (\S\ref{sec:annot_blockmerge}) that defines the unit every later measurement is computed over.

\subsection{Segmentation}
\label{sec:annot_seg}
\looseness=-1
Counselling e-mails are long and weave several concerns and emotions together.
A counsellor message greets, analyses, offers help and signs off.
A client message runs through several emotions in turn.
So a single label per message is rarely meaningful.
To keep this internal structure, the pipeline first splits each e-mail into spans (stage~1, Fig.~\ref{fig:pipeline}), semantic blocks that usually coincide with a sentence, each carrying one counsellor act and one emotion label.
Counselling text often runs on without clear sentence boundaries, so each e-mail is segmented with the pre-trained Segment-any-Text (SaT) model \cite{frohmann2024sat}, a neural segmenter that recovers boundaries even where punctuation is missing.
On inspection this is more reliable than splitting on punctuation alone.
The segmentation is kept deliberately fine.
A per-taxonomy block-merge (\S\ref{sec:annot_blockmerge}) recombines same-label neighbours afterwards, so over-splitting is undone while a coarser split would risk collapsing distinct labels into one.
GEMCo-A contributes 9{,}444 spans and GEMCo-B 4{,}493 (13{,}937 proxy), the real data a further 14{,}388.
Together that is 28{,}325 spans across the 1{,}315 messages of proxy and the real data combined.

\subsection{Counsellor Acts (OnCoCo)}
\label{sec:annot_oncoco}
Counsellor spans are classified with OnCoCo \cite{oncoco2026} (stage~2, Fig.~\ref{fig:pipeline}), a counselling-act taxonomy built for online counselling, with the dataset and classifier publicly released.
Its cross-validated macro $F_1$ of $.72$ is measured at the finest leaf level and rises to $.83$ when the right label is among the top two.
The remaining errors mostly fall on neighbouring categories.
It was trained on a bilingual German--English corpus of counselling messages manually segmented into labelled act spans, the same span-level input SaT produces, so its training and application share one segmentation.
Being bilingual, it also labels the English cross-domain corpora (Appendix~\ref{app:baselines}).
This paper collapses its output to the nine top-level categories, each capturing what the counsellor does (Figure~\ref{fig:strategies}), coarser and correspondingly more reliable than the leaf level.
Four are formal: formalities opening [FA] and closing [FC] a message, moderation [Mod] that structures the reply and other [O].
Five are therapeutic-process \textit{Impact Factors}: analysis \& clarification [AC], agreement on objectives [AO], creating motivation [CM], resource activation [RA] and help \& problem solving [HP].
Per-category descriptions with examples are in Appendix~\ref{app:categories}.

\begin{figure*}[t]
    \centering
    \includegraphics[width=0.93\textwidth]{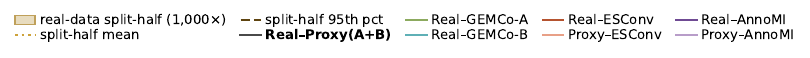}\\[1pt]
    \begin{subfigure}[t]{0.49\textwidth}
        \centering
        \includegraphics[width=\linewidth]{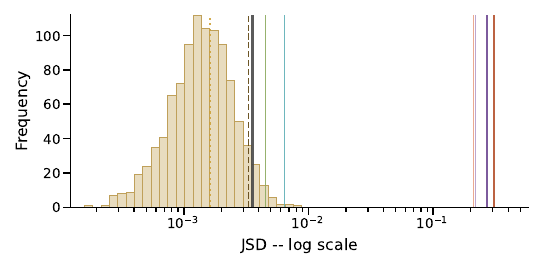}
        \caption{Counsellor strategy (OnCoCo, 9 cat.).}
        \label{fig:splithalf_a}
    \end{subfigure}\hfill
    \begin{subfigure}[t]{0.49\textwidth}
        \centering
        \includegraphics[width=\linewidth]{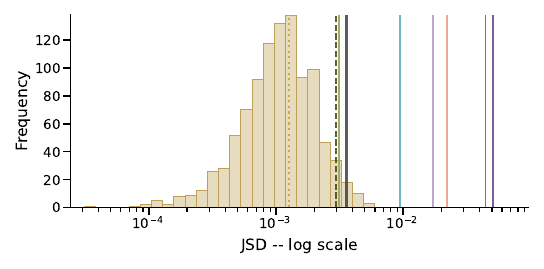}
        \caption{Client emotion (Ekman, 7 cat.).}
        \label{fig:splithalf_b}
    \end{subfigure}
    \caption{Two-baseline scale for the proxy--real JSD (log axis).}
    \label{fig:splithalf}
\end{figure*}

\subsection{Client Emotions (Ekman)}
\label{sec:annot_emotion}
Each client span is additionally classified for emotion (stage~2, Fig.~\ref{fig:pipeline}) with the classifier of \citet{lalk2025emotion}, which predicts the 28 GoEmotions categories \cite{demszky2020goemotions}.
These 28 categories are mapped onto Ekman's six basic emotions plus neutral \cite{ekman1992basic}, the seven labels this analysis uses for what the client feels, following the mapping GoEmotions itself defines.
That mapping table and the fuller GoEmotions-level analysis are given in Appendix~\ref{app:goemotions}.
The classifier is the closest available fit for GEMCo, developed for emotion detection in psychotherapy transcripts and fine-tuned on German.
It reaches $F_1 = .45$ over all 28 categories, on par with the original English GoEmotions benchmark ($.46$).
Being multilingual, it also labels the English cross-domain corpora (Appendix~\ref{app:baselines}).
Collapsing to the seven Ekman classes lowers the resolution but folds together several categories the classifier tends to confuse.

\subsection{Per-Taxonomy Block-Merging}
\label{sec:annot_blockmerge}
SaT segments text more finely than the labels actually change.
A single problem description can run over several semantic blocks that carry the same act or emotion and counting each separately would over-weight long passages.
Before any analysis, consecutive spans with the same label are therefore merged into one block (stage~3, Fig.~\ref{fig:pipeline}).
For counsellor strategy, ``Hello,'' and ``how are you doing?'' are both \textit{Formalities (beginning)} and become one block.
On the client side, ``I cry all the time.'' and ``Nothing brings me joy anymore.'' are both \textit{sadness} and merge likewise.
Each message then reads as a clean sequence of distinct OnCoCo acts on the counsellor side and of distinct emotions on the client side.

\section{Corpus-Level Validation}
\label{sec:corpus_validation}
\label{sec:methodology}

Proxy is validated against the real data first by a vocabulary check (Section~\ref{sec:lexical}), then through two label-based lenses, the counsellor through \textit{strategy} (OnCoCo acts, what they do) and the client through \textit{emotion} (Ekman, what they feel).

Each lens pools all the labelled blocks of a corpus into one distribution and compares proxy against the real data with a single corpus-level number, the Jensen--Shannon divergence (JSD), which runs in $\log_2$ from $0$ for identical distributions to $1$ when they share no category (full definition, Appendix~\ref{app:stat_method}).

Comparing the real data to a second sample of itself adapts classic split-half reliability \cite{spearman1910,brown1910}, while comparing the constructed proxy corpus to the real one follows distribution-level evaluation of model-generated against real data \cite{esteban2017realvalued,pillutla2021mauve}.


\subsection{Vocabulary Overlap}
\label{sec:lexical}

Before any labels, a classifier-free question: do the proxy and the real data draw on the same words?
The measure is the Jaccard overlap, the share of word types two corpora have in common.
The baseline is the real data against itself. Its 124 conversations are split into two random halves and their Jaccard overlap is measured. Repeated over $1{,}000$ random splits, this gives a tight reference band (Jaccard $J^{\mathrm{RR}} = \jaccSplit \pm \jaccSplitSD$).
The superscript names the two corpora compared, the real data first, so $\mathrm{RR}$ is the real data against a second sample of itself and $\mathrm{RP}$ is the real data against the proxy.
By speaker, the counsellor side reaches its own band ($J^{\mathrm{RP}}_{\mathrm{counsellor}} = \jaccRPCo$ vs.\ $J^{\mathrm{RR}}_{\mathrm{counsellor}} = \jaccSplitCo \pm \jaccSplitCoSD$), so the professional wording is about as close as the real data is to itself, while the client side stays below it ($J^{\mathrm{RP}}_{\mathrm{client}} = \jaccRPCl$ vs.\ $J^{\mathrm{RR}}_{\mathrm{client}} = \jaccSplitCl \pm \jaccSplitClSD$).

\subsection{Counsellor Strategy Comparison}
\label{sec:strategies}

Vocabulary overlap compares the words.
The counsellor strategy check compares the counsellor acts, asking whether proxy deploys the same professional repertoire as the real counselling (full resolution in Appendix~\ref{app:full_dists}).
The gap is read on a scale with two empirical ends (Figure~\ref{fig:splithalf_a}, Table~\ref{tab:jsd_main}).
The low end is the real data against itself.
The same $1{,}000$ half-splits scored by JSD form its own \emph{noise band}, with the mean as the noise floor ($\mathrm{JSD}^{\mathrm{RR}}_{\mathrm{OnCoCo}}$) and the 95th percentile ($\mathrm{P95}^{\mathrm{RR}}$) as the upper edge.
The high end is the real data against the two external counselling corpora, far higher at $.31$ and $.27$.
Proxy's gap ($\mathrm{JSD}^{\mathrm{RP}}_{\mathrm{OnCoCo}}$, Table~\ref{tab:jsd_main}) sits just past that upper edge, where \shPctlCo\,\% of the within-real splits fall below it, with a small-to-medium effect size (Cram\'er's~$V$, Appendix~\ref{app:stat_method}).
The gap is small but systematic, measuring $.020$ on the coarser native segmentation, still an order of magnitude below the external anchors.

\subsection{Client Emotion Comparison}
\label{sec:emotions}
The strategy check compares the counsellor's acts.
Emotion does this for the client, asking whether proxy's clients move through the same feelings as the real ones, on the same two-ended scale (Figure~\ref{fig:splithalf_b}).
The pooled proxy--real gap ($\mathrm{JSD}^{\mathrm{RP}}_{\mathrm{Ekman}}$, Table~\ref{tab:jsd_main}) sits at the \shPctlEk th~percentile of the within-real noise band, near its edge.
On the coarser native segmentation it sits two-to-three-fold below the external corpora ($.017$ vs.\ $.044$ and $.051$, Appendix~\ref{app:baselines}).
The per-subcorpus rates (Appendix~\ref{app:full_dists}) locate two shifts inside this small gap.
A difference is counted only where a subcorpus's 95\,\% confidence interval lies fully outside the real data’s.
GEMCo-A reads a little warmer (more \textit{joy}, $\sim$30\,\% vs.\ $\sim$25\,\%) but stays inside that interval, while GEMCo-B's extra \textit{surprise} falls outside it, its role-played clients overplaying their assigned cases.
The professional side stays inside the noise band and the constructed client carries the difference.

\begin{table}[H]
\centering
\footnotesize
\begin{tabular*}{\columnwidth}{@{\extracolsep{\fill}}lcccc@{}}
\toprule
\textbf{Taxonomy} & $\mathrm{JSD}^{\mathrm{RP}}$ & \textbf{95\,\% CI} & $\boldsymbol{V}$ & $\mathrm{JSD}^{\mathrm{RR}}$ \\
\midrule
OnCoCo (9 cat.) & \shRPCo & \ciCoLo--\ciCoHi & \vCo & \shRRCo \\
Ekman (7 cat.) & \shRPEk & \ciEkLo--\ciEkHi & \vEk & \shRREk \\
\bottomrule
\end{tabular*}
\caption{Corpus-level proxy--real gap ($\mathrm{JSD}^{\mathrm{RP}}$) against the within-real noise floor (split mean $\mathrm{JSD}^{\mathrm{RR}}$); percentiles in the text, size-matched null in Appendix~\ref{app:splithalf}.}
\label{tab:jsd_main}
\end{table}

A machine-translated ESConv or AnnoMI would not substitute for GEMCo.
Their distance to the real data ($.27$--$.31$ vs.\ the proxy's $.020$) is counselling format, modality and register along with language.
Translation removes only the language part.
The emotion classifier adds a caution: trained on DeepL-translated GoEmotions, it shows a macro-$F_1$ gap across the language boundary ($.45$ German vs.\ $.38$ English, Appendix~\ref{app:baselines}).

\section{Conversation-Progress Validation}
\label{sec:bins}

Conversation progress is the next check.
It confirms that the match holds along the thread and that no opposite drifts cancel in the corpus.
Each conversation is cut into five equal progress bins and read one by one, for GEMCo-A, GEMCo-B and the pooled corpus (Section~\ref{sec:per_category} traces where the two subcorpora differ).
Progress position is only well defined for a thread that runs to completion, so this view uses the 48 complete real threads with at least five messages, the same reference as the transitions in Section~\ref{sec:dyad}. The all-124 version is similar and given in Appendix~\ref{app:splithalf}.

\subsection{Counsellor Strategy per Bin}
\label{sec:bins_co}

\begin{figure}[!ht]
    \centering
    \includegraphics[width=\columnwidth]{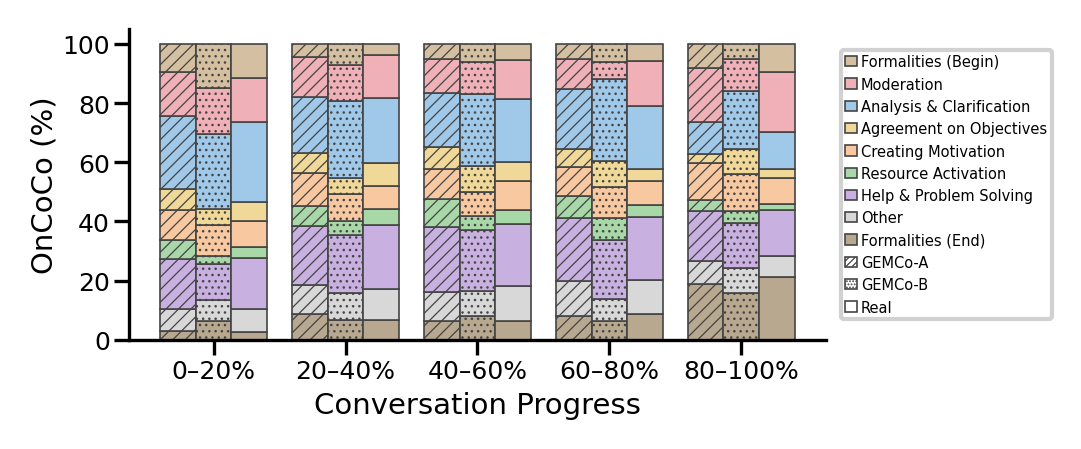}
    \caption{Counsellor strategy (OnCoCo, 9 cat.) over conversation progress.
GEMCo-A hatched, GEMCo-B dotted, Real plain.}
    \label{fig:strategies}
\end{figure}

\looseness=-1
The nine strategy categories move together across GEMCo-A, GEMCo-B and the real data, steady along the thread (Figure~\ref{fig:strategies}, Table~\ref{tab:jsd_perbin}).
The shared bend is \textit{Analysis~\& Clarification} tapering toward the end, the arc from understanding a case to resolving it.

Pooling the two subcorpora helps on the counsellor side.
The pooled GEMCo gap lands below both GEMCo-A and GEMCo-B at the corpus level and through the first three bins.
The two lean off the real data in different directions and partly cancel.
GEMCo-A also stays inside the real data’s own noise band in every bin.
A per-bin value in bold in Table~\ref{tab:jsd_perbin} lies above that band's 95th percentile ($\mathrm{P95}^{\mathrm{RR}}$).
GEMCo-B is the one that breaks the band, in the closing two bins.
Its role-played sessions still work the case to the end and lift \textit{Analysis~\& Clarification} where the real sessions wind down.
Even there the gap stays at least five-fold below the cross-domain distance ($.27$--$.31$, Table~\ref{tab:jsd_perbin}).

\subsection{Client Emotion per Bin}
\label{sec:bins_cl}

\begin{figure}[!ht]
    \centering
    \includegraphics[width=\columnwidth]{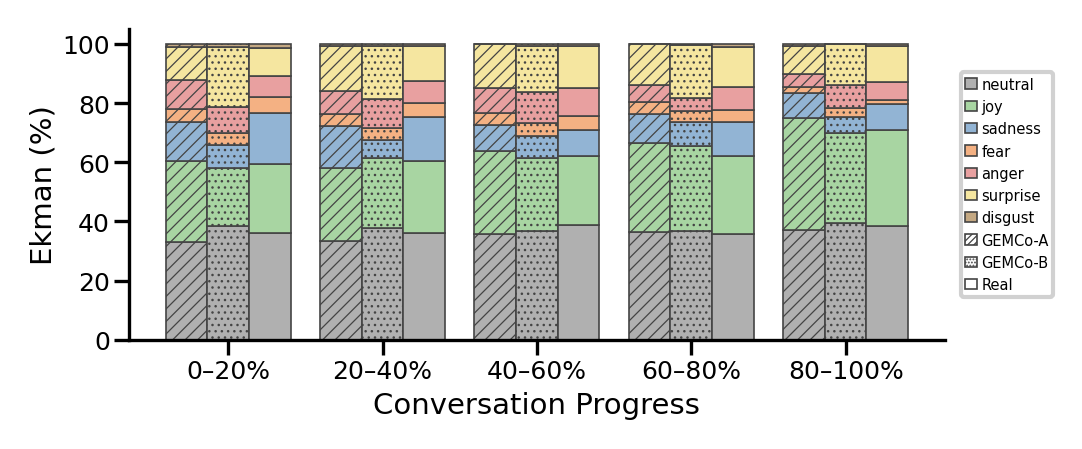}
    \caption{Client emotion (Ekman, 7 cat.) over conversation progress.
GEMCo-A hatched, GEMCo-B dotted, Real plain.}
    \label{fig:emotions}
\end{figure}

Client emotion follows a clearer arc (Figure~\ref{fig:emotions}).
In the real data and GEMCo-A, clients begin with elevated \textit{sadness}, \textit{fear} and \textit{anger} that diminish as \textit{joy} rises, the course a counselling exchange works toward.
This is a structural signature, not direct evidence of a working alliance.
Here GEMCo-A alone is closest to the real data and the pooled gap sits a little above it.
GEMCo-B's role-played clients overplay the emotion (Table~\ref{tab:jsd_perbin}).
GEMCo-A holds inside the band throughout, only a touch warmer.
GEMCo-B opens outside it in the first two bins.
Its clients lead with \textit{surprise} in place of the early \textit{sadness} of real threads.
The pooled corpus still stays inside the band, so adding GEMCo-B widens the case diversity at a small cost in emotional fit.

\begin{table}[t]
\centering
\scriptsize
\setlength{\tabcolsep}{2.5pt}
\begin{tabular*}{\columnwidth}{@{\extracolsep{\fill}}l ccc ccc@{}}
\toprule
& \multicolumn{3}{c}{\textbf{Strategy}} & \multicolumn{3}{c}{\textbf{Emotion}} \\
\cmidrule(lr){2-4} \cmidrule(lr){5-7}
\textbf{Progress} & A & B & GEMCo & A & B & GEMCo \\
\midrule
\perbinSplitRows
\midrule
cross-domain & \multicolumn{3}{c}{.27--.31} & \multicolumn{3}{c}{.04--.05} \\
\bottomrule
\end{tabular*}
\caption{Per-bin $\mathrm{JSD}$ against the 48 complete real threads, for GEMCo-A (A), GEMCo-B (B) and pooled \textbf{GEMCo}.}
\label{tab:jsd_perbin}
\end{table}

\section{Within-Message Transitions}
\label{sec:dyad}

Sections~\ref{sec:corpus_validation} and~\ref{sec:bins} checked how often each strategy and emotion occurs.
This section checks their order, which act or emotion follows which inside a message, for both the counsellor and the client.
Such transition structure also supports next-act prediction in counselling dialogue~\cite{rudolph2026transition}.
Each transition meets the same 95th-percentile test as the corpus gap (\S\ref{sec:strategies}), now per transition, so each gets its own noise band.
The false positives across the many tests are bounded by FDR control (Appendix~\ref{app:fdr}).
The reference is the 48 real threads with at least five messages, where a full back-and-forth develops.
Proxy holds only such complete threads.
Full method and matrices are in Appendix~\ref{app:within_co}.

\subsection{Counsellor Strategy Stability}
The counsellor transitions stay within the real data’s noise.
After correcting for the many transitions tested (FDR), a single shift stays significant: in GEMCo-B, fewer openings go straight to a sign-off (\textit{Formalities (beginning)}$\to$\textit{Formalities (conclusion)}, $-7.6$\,pp).
It does not recur in GEMCo-A.
With band medians of only $\tauMedCoA$ and $\tauMedCoB$\,pp, a consistent shift would have surfaced, so within the message the proxy's professionals order their strategy as the real data does.

\subsection{Client Emotion Shifts}
\looseness=-1
The client side does shift.
The FDR survivors (\nFdrEkA{} in GEMCo-A and \nFdrEkB{} in GEMCo-B) show the same positive-affect signature already seen in the distributions (\S\ref{sec:emotions}).
GEMCo-A lifts \textit{surprise}$\to$\textit{joy} ($+2.0$\,pp).
GEMCo-B gains \textit{neutral}$\leftrightarrow$\textit{surprise} and loses \textit{sadness} (\textit{neutral}$\to$\textit{sadness}, $-3.1$\,pp).
The difference is in the order of feelings, not only their rates.

Together the two lenses place the residual divergence on the constructed client rather than on how the real counsellors sequence their work.

\section{Cross-Speaker Interplay}
\label{sec:interplay}

\looseness=-1
The checks so far read each speaker on their own.
This section puts the two together and asks whether counsellor and client fit each other in the proxy as they do in the real data.
Linguistic Style Matching (LSM) \cite{niederhoffer2002linguistic} measures whether the two speakers settle into the same way of writing, comparing their use of eight closed-class German function-word categories, topic-free words whose matching shows two people attuning.
LSM uses pooled proxy and all 124 real threads, with no need for a thread to run to completion.
Each conversation gets one score from $0$ to $1$.
Both corpora score high (medians $\lsmProxyMed$ proxy, $\lsmRealMed$ real), a touch higher in the real data.
A Kolmogorov--Smirnov test \cite{massey1951kolmogorov} puts the distance between the distributions at $D = \lsmKS$ and resampling the real data against itself gives a noise band whose 95th percentile is $\lsmNullPNF$, which $D$ exceeds.
Proxy pairs thus align slightly less tightly than real ones.
Counsellor and client attune more in the real exchange than in drafted GEMCo-A or role-played GEMCo-B.
The German function words have no English equivalent, so this check has no cross-domain anchor.

\section{Where Proxy and the Real Data Differ}
\label{sec:per_category}

Sections~\ref{sec:corpus_validation}--\ref{sec:interplay} established that proxy tracks the real data closely.
The residual differences fall primarily on the authored (GEMCo-A) and role-played (GEMCo-B) clients.
Table~\ref{tab:per_cat} locates them, the three categories (of the nine strategies and seven Ekman emotions) where GEMCo-B's 95\,\% CI lies fully outside the real data’s, no GEMCo-A category clearing it (Appendix~\ref{app:full_dists}).

\begin{table}[H]
\centering
\footnotesize
\setlength{\tabcolsep}{3pt}
\resizebox{\columnwidth}{!}{%
\begin{tabular}{@{}lrrrc@{}}
\toprule
\textbf{Category} & \textbf{GEMCo-A} & \textbf{GEMCo-B} & \textbf{Real} & \textbf{Pattern} \\
\midrule
\multicolumn{5}{l}{\textit{Counsellor strategies}} \\
\quad Moderation          & 13.7\,\textpm\,2.0 & 11.0\,\textpm\,2.3 & 16.1\,\textpm\,1.5 & B\,$\downarrow$ \\
\midrule
\multicolumn{5}{l}{\textit{Client emotions (Ekman)}} \\
\quad surprise            & 12.9\,\textpm\,2.3 & 17.1\,\textpm\,3.2 & 12.1\,\textpm\,1.5 & B\,$\uparrow$ \\
\quad sadness             & 11.0\,\textpm\,2.1 & 6.9\,\textpm\,2.2  & 12.7\,\textpm\,1.6 & B\,$\downarrow$ \\
\bottomrule
\end{tabular}}
\caption{Per-subcorpus \% for the three categories where GEMCo-B's CI lies fully outside the real data’s.
\textit{Pattern}: B = GEMCo-B, arrow = direction.}
\label{tab:per_cat}
\end{table}

\subsection{Counsellor-Side Construction Effects}
On the counsellor side the construction barely shows.
Vocabulary, the corpus act gap ($\mathrm{JSD}^{\mathrm{RP}}_{\mathrm{OnCoCo}} = \shRPCo$), the per-bin arc and the within-message order all stay inside the real data’s band (\S\ref{sec:lexical}--\S\ref{sec:dyad}), only GEMCo-B's lower \textit{Moderation} clearing its CI.
With little genuine back-and-forth to steer, less moderation is needed and motivational moves fill the space, so the residual sits on the client side.

\subsection{Two Client Production Modes}
On the client side the corpus emotion gap stays tiny ($\mathrm{JSD}^{\mathrm{RP}}_{\mathrm{Ekman}} = \shRPEk$), GEMCo-A leaning to more \textit{joy} and GEMCo-B to more \textit{surprise}, GEMCo-B's opening bins and the order of feelings carrying it across corpus, thread and message (\S\ref{sec:lexical}--\S\ref{sec:interplay}).
These trace to two production modes.
GEMCo-A's expert authors front-load warmth and motivation, a \emph{narrative-closure} mode whose elevated \textit{joy} stays within the real data’s CI.
GEMCo-B's role-players show the complementary \emph{problem-fixation} mode, restating the assigned case where the real data lets the solution play out.
The elevated \textit{surprise} ($+5$\,pp) and reduced \textit{sadness} ($-6$\,pp) are the shifts that clear the CI.

\subsection{The Paired-Design Advantage}
GEMCo-A and GEMCo-B are complementary and lean off the real data in slightly different directions.
Pooling lets their biases partly cancel, so the combined proxy lands closer to the real data than their average.
The pooled strategy gap ($\mathrm{JSD}^{\mathrm{RP}}_{\mathrm{OnCoCo}} = \shRPCo$; Figure~\ref{fig:splithalf_a}) falls below both GEMCo-A's ($.0045$) and GEMCo-B's ($.0064$); for emotion the pool ($\shRPEk$) beats their average but not GEMCo-A alone ($.0031$).
Pooling also widens the diversity of cases and counsellor styles and yields more genuine counsellor messages, a richer training signal, so the two are best used together.

Distribution, order and cross-speaker interplay all hold up, but they read labels.
The next section therefore puts the pooled corpus to a generative test, where using the two together pays off (Section~\ref{sec:downstream}).

\section{Generative Validation}
\label{sec:downstream}
\setlength{\intextsep}{4pt plus 1pt minus 1pt}
\setlength{\abovecaptionskip}{3pt}

The earlier sections show that proxy and real match in what they contain.
This last check asks whether the proxy also works as training data: a model trained only on it should write counsellor replies closer to the real ones than the off-the-shelf model does.
The probe tests the corpus as a training signal, not a counselling generator to deploy.
Mistral-Small-3.2-24B-Instruct \cite{mistral2025small} was fine-tuned on the proxy counsellor turns alone, never on the real data, yielding \textit{Proxy-SFT} (QLoRA adapter, Appendix~\ref{app:genval}).
At every counsellor turn of the \dsNreal{} held-out real threads, three candidates write the next reply: the human counsellor, Proxy-SFT and the off-the-shelf base model with a strong counsellor prompt.
Each is conditioned on the real conversation up to that turn, so the three differ only in the one reply.

\subsection{LLM-as-Judge Ranking}
A blind Mistral-Medium-3.5 model \cite{zheng2023judging} judged the candidates pairwise, each item showing the real conversation up to that point and two replies (A and B) and asking which better fits as the next counsellor message.
Positions were swapped for every sample (AB vs.\ BA) to cancel any position bias (\dsNjudg{} judgements, Table~\ref{tab:genval_rank}; prompt in Appendix~\ref{app:judge_prompt}).
The judge ran self-hosted, so no conversation data left the institution.

\begin{table}[H]
\centering
\small
\begin{tabular*}{\columnwidth}{@{\extracolsep{\fill}}lccc@{}}
\toprule
System & WR\,\% [95\,\% CI] & BT & Rank \\
\midrule
\textbf{Human (reference)} & \dsHumanWR\ [\dsHumanWRlo--\dsHumanWRhi] & \dsBThuman{} & 1 \\
Proxy-SFT & \dsProxyWR\ [\dsProxyWRlo--\dsProxyWRhi] & \dsBTproxy{} & 2 \\
Off-the-shelf instruct & \dsInstructWR\ [\dsInstructWRlo--\dsInstructWRhi] & \dsBTinstruct{} & 3 \\
\bottomrule
\end{tabular*}
\caption{Win rate (WR, bootstrap 95\,\% CI), Bradley--Terry strength (BT) and rank per system.}
\label{tab:genval_rank}
\end{table}

The Bradley--Terry strength \cite{bradley1952rank} condenses the pairwise outcomes into one score per system, its ratios giving the head-to-head odds.
The human is preferred to both models, head to head \dsHumanBeatsProxy{}:\the\numexpr100-\dsHumanBeatsProxy\relax{} over Proxy-SFT and \dsHumanBeatsInstruct{}:\the\numexpr100-\dsHumanBeatsInstruct\relax{} over the off-the-shelf model.
The proxy-tuned model beats the off-the-shelf one \dsProxyBeatsInstruct{}:\the\numexpr100-\dsProxyBeatsInstruct\relax{} and sits closer to the human.
The proxy adapter thus appears to move output toward authentic counsellor messages.
The judge is stable under order swaps (AB/BA agreement \dsConsistency\,\%) and the shared Mistral-Small-3.2-24B backbone would favour the two candidates equally, not the non-Mistral human who still ranks first (Appendix~\ref{app:genval}).

\subsection{Embedding-Based Check: MAUVE}
The judge states preferences.
MAUVE \cite{pillutla2021mauve} asks instead whether the model's replies occupy the same distribution as the human ones.
Every reply is embedded on its own (message text only, no context) as one vector from \texttt{Ministral-8B-Instruct-2410} \cite{mistral2024ministral} (last-token hidden state).
MAUVE summarises the divergence between the two embedding clouds on a $0$--$1$ scale, $1$ = indistinguishable, $0$ = no common region.
Each row of Table~\ref{tab:genval_mauve} compares two disjoint halves of the 124 real conversations, matching one half's human replies against the other half's (the ceiling), Proxy-SFT's and the off-the-shelf model's, size-matched over 50 splits (Appendix~\ref{app:mauve}).

\begin{table}[H]
\centering
\small
\begin{tabular*}{\columnwidth}{@{\extracolsep{\fill}}lc@{}}
\toprule
Comparison & MAUVE \\
\midrule
Human $\leftrightarrow$ Human (split-half) & $\mauveHH \pm \mauveHHsd$ \\
Human $\leftrightarrow$ Proxy-SFT & $\mauveXSft \pm \mauveXSftSd$ \\
Human $\leftrightarrow$ Off-the-shelf instruct & $\mauveXInstruct \pm \mauveXInstructSd$ \\
\bottomrule
\end{tabular*}
\caption{MAUVE over 50 conversation-level splits.}
\label{tab:genval_mauve}
\end{table}

The Proxy-SFT replies sit close to the within-human ceiling, within about a standard deviation of it.
The off-the-shelf model, given the same conversation prefixes and a strong counsellor prompt, reaches barely a quarter of the scale.
The distance between the two reflects what the adapter learned from proxy's counsellor side, the side the validation placed inside the real data’s noise band.

\section{Summary}
\label{sec:discussion}

GEMCo is a shareable, asynchronous counselling corpus that complies with research ethics and stays close to the withheld real data on every measure tested here.
The proxy--real gap sits just above the real data’s own noise floor and far below the cross-domain distances. Proxy reproduces what counsellors and clients do, without copying how the real data is written.

The small remaining gap falls on the constructed client side, arguably the more tolerable one.
The dyadic style is slightly looser and the register lighter while the professional side stays closest to the real data.

GEMCo-A and GEMCo-B are complementary, their writing modes partly cancelling when pooled (\S\ref{sec:per_category}), so the paired corpus approximates the real data better than their average.
The same design could transfer to other low-resource, ethically constrained domains, pairing a small withheld reference with releasable human data read against its own noise band.

\section*{Limitations}

GEMCo is small, 86 threads, the scale of AnnoMI and HOPE.
Further field deployments of the GEMCo-B type~\cite{steigerwald2025caia} can extend it.
The intended use is evaluation and fine-tuning rather than pretraining and Section~\ref{sec:downstream} shows the 86 threads already carry a usable training signal.
The open release lets third parties add annotation layers and pool GEMCo with other German counselling data.
Its validation is distributional and sequential, read through classifiers, showing no clinical or therapeutic equivalence.
It shows that proxy deploys the professional repertoire at the real data’s rates and in the real data’s order, not that any single reply is good counselling.
The JSD is dominated by the common categories, so a rare category can differ sharply between proxy and the real data while barely changing the score.
Equivalence on such rare categories is therefore not claimed.
To offset this, the conversation-progress bins (\S\ref{sec:bins}) and within-message transitions (\S\ref{sec:dyad}) test the match at a finer grain and in sequence.
It also assumes the classifiers make the same systematic errors on proxy as on the real data (\S\ref{sec:annot_emotion}), plausible but unverified.
Both corpora pass through the same instrument, so a systematic classifier bias shifts both distributions together and largely cancels in the comparison.
What does not cancel is per-category reliability, with emotion ($F_1 = .45$) less reliable than strategy ($.72$).
Collapsing emotion to the seven Ekman classes should absorb some of that error, though the gain is unverified.
Every label here is classifier-produced rather than human-annotated, so a human audit of a sample is the natural next step.
Higher-level counselling qualities are not measured: empathy, therapeutic appropriateness, coherence and conversational effectiveness.
Validated empathy scales and counsellor ratings could measure them.
The generative check relies on an LLM judge and MAUVE, with no rating by professional counsellors.
A counsellor-rated comparison of generated replies would give stronger evidence.
GEMCo also covers only a few counselling areas (\S\ref{sec:corpus}), such as youth, family and addiction counselling, while counselling itself can address almost any topic, so the corpus generalises only so far.

\section*{Intended Uses and Ethics}
\label{sec:ethics}

GEMCo (the proxy) is meant to open counselling support for German NLP research, not to automate care.
The expected primary use is counsellor-facing.
The tools it supports stay human-in-the-loop: drafting and supervision aids, training, quality assurance and triage where professionals are scarce~\cite{steigerwald2025ethics}.
GEMCo itself still needs further preparation for such tools, but it could help widen access to mental-health support by pairing AI with human counsellors.

Several uses fall outside what the corpus supports: unsupervised counsellor chatbots, crisis response without human escalation and diagnosis or clinical decision support.
There is no outcome data for them and the dual-use risk is clear.

GEMCo-A is fully fictional. GEMCo-B comes from a study whose participants were all fully informed and consented to publication. The real data was donated under informed consent, revocable at any time (\S\ref{sec:corpus}).
The real data is never released and only aggregate statistics of it appear in this paper. The generative probe trains only on proxy, which holds no real client data.

\section*{Acknowledgments}
Funded by the Deutsche Forschungsgemeinschaft (DFG, German Research Foundation) -- FIP 160 -- Project-ID 549142762.

\bibliography{references}

@article{malgaroli2023nlp,
  title={Natural Language Processing for Mental Health Interventions: A Systematic Review and Research Framework},
  author={Malgaroli, Matteo and Hull, Thomas D and Zech, James M and Althoff, Tim},
  journal={Translational Psychiatry},
  volume={13},
  number={1},
  pages={309},
  year={2023},
  publisher={Nature Publishing Group},
  doi={10.1038/s41398-023-02592-2}
}

@article{ekman1992basic,
  title={An Argument for Basic Emotions},
  author={Ekman, Paul},
  journal={Cognition and Emotion},
  volume={6},
  number={3--4},
  pages={169--200},
  year={1992},
  doi={10.1080/02699939208411068}
}

@inproceedings{demszky2020goemotions,
  title={{GoEmotions}: A Dataset of Fine-Grained Emotions},
  author={Demszky, Dorottya and Movshovitz-Attias, Dana and Ko, Jeongwoo and Cowen, Alan and Nemade, Gaurav and Ravi, Sujith},
  booktitle={Proceedings of the 58th Annual Meeting of the Association for Computational Linguistics},
  pages={4040--4054},
  year={2020},
  doi={10.18653/v1/2020.acl-main.372}
}

@article{lalk2025emotion,
  title={Employing large language models for emotion detection in psychotherapy transcripts},
  author={Lalk, Christopher and Targan, Kim and Steinbrenner, Tobias and Schaffrath, Jana and Eberhardt, Steffen and Schwartz, Brian and Vehlen, Antonia and Lutz, Wolfgang and Rubel, Julian},
  journal={Frontiers in Psychiatry},
  volume={16},
  year={2025},
  doi={10.3389/fpsyt.2025.1504306}
}

@inproceedings{steigerwald2025ethics,
  title={{AI} Systems in Text-Based Online Counselling: Ethical Considerations Across Three Implementation Approaches},
  author={Steigerwald, Philipp and Burghardt, Jennifer and Rudolph, Eric and Albrecht, Jens},
  booktitle={Proceedings of the 3rd International Conference on Frontiers of Artificial Intelligence, Ethics, and Multidisciplinary Applications (FAIEMA)},
  year={2025}
}

@inproceedings{steigerwald2025caia,
  title={{CAIA} in Practice: Field Evaluation of an {AI}-Assisted Support System for Text-Based Online Counselling},
  author={Steigerwald, Philipp and Bienlein, Nico and Burghardt, Jennifer and Stieler, Mara and Lehmann, Robert and Albrecht, Jens},
  booktitle={Proceedings of the 37th IEEE International Conference on Tools with Artificial Intelligence (ICTAI)},
  doi={10.1109/ICTAI66417.2025.00214},
  year={2025}
}

@inproceedings{rudolph2025roleplay,
  title={Comparing Human Role-Players and {LLM}-Simulated Clients in Online Counselling Training: An Analysis of Counselling Patterns},
  author={Rudolph, Eric and Steigerwald, Philipp and Albrecht, Jens},
  booktitle={Proceedings of the 18th International Conference on Educational Data Mining (EDM)},
  pages={381--387},
  doi={10.5281/zenodo.15870260},
  year={2025}
}

@inproceedings{rudolph2026transition,
  title={Transition-Matrix Regularization for Next Dialogue Act Prediction in Counselling Conversations},
  author={Rudolph, Eric and Steigerwald, Philipp and Albrecht, Jens},
  booktitle={Findings of the Association for Computational Linguistics: ACL 2026},
  pages={25442--25457},
  doi={10.18653/v1/2026.findings-acl.1271},
  year={2026}
}

@inproceedings{liu2021ESConv,
  title={Towards Emotional Support Dialog Systems},
  author={Liu, Siyang and Zheng, Chujie and Demasi, Orianna and Sabour, Sahand and Li, Yu and Yu, Zhou and Jiang, Yong and Huang, Minlie},
  booktitle={Proceedings of the 59th Annual Meeting of the Association for Computational Linguistics and the 11th International Joint Conference on Natural Language Processing},
  pages={3469--3483},
  year={2021},
  doi={10.18653/v1/2021.acl-long.269}
}

@article{althoff2016counseling,
  title={Large-scale Analysis of Counseling Conversations: An Application of Natural Language Processing to Mental Health},
  author={Althoff, Tim and Clark, Kevin and Leskovec, Jure},
  journal={Transactions of the Association for Computational Linguistics},
  volume={4},
  pages={463--476},
  year={2016},
  doi={10.1162/tacl_a_00111}
}

@inproceedings{gratch2014daic,
  title={The Distress Analysis Interview Corpus of Human and Computer Interviews},
  author={Gratch, Jonathan and Artstein, Ron and Lucas, Gale and Stratou, Giota and Scherer, Stefan and Nazarian, Angela and Wood, Rachel and Boberg, Jill and DeVault, David and Marsella, Stacy and Traum, David and Rizzo, Skip and Morency, Louis-Philippe},
  booktitle={Proceedings of the Ninth International Conference on Language Resources and Evaluation ({LREC}'14)},
  pages={3123--3128},
  year={2014},
  address={Reykjavik, Iceland},
  publisher={European Language Resources Association (ELRA)}
}

@inproceedings{wu2022annomi,
  title={{Anno-MI}: A Dataset of Expert-Annotated Counselling Dialogues},
  author={Wu, Zixiu and Balloccu, Simone and Kumar, Vivek and Helaoui, Rim and Reiter, Ehud and Reforgiato Recupero, Diego and Riboni, Daniele},
  booktitle={Proceedings of {ICASSP} 2022 -- {IEEE} International Conference on Acoustics, Speech and Signal Processing},
  pages={6177--6181},
  year={2022},
  doi={10.1109/ICASSP43922.2022.9746035}
}

@inproceedings{zanwar2023smhdger,
  title={{SMHD-GER}: A Large-Scale Benchmark Dataset for Automatic Mental Health Detection from Social Media in {G}erman},
  author={Zanwar, Sourabh and Wiechmann, Daniel and Qiao, Yu and Kerz, Elma},
  booktitle={Findings of the Association for Computational Linguistics: EACL 2023},
  pages={1526--1541},
  year={2023},
  doi={10.18653/v1/2023.findings-eacl.113}
}

@inproceedings{malhotra2022hope,
  title={Speaker and Time-aware Joint Contextual Learning for Dialogue-act Classification in Counselling Conversations},
  author={Malhotra, Ganeshan and Waheed, Abdul and Srivastava, Aseem and Akhtar, Md Shad and Chakraborty, Tanmoy},
  booktitle={Proceedings of the Fifteenth ACM International Conference on Web Search and Data Mining (WSDM)},
  pages={735--745},
  year={2022},
  doi={10.1145/3488560.3498509}
}

@inproceedings{sun2021psyqa,
  title={{PsyQA}: A {C}hinese Dataset for Generating Long Counseling Text for Mental Health Support},
  author={Sun, Hao and Lin, Zhenru and Zheng, Chujie and Liu, Siyang and Huang, Minlie},
  booktitle={Findings of the Association for Computational Linguistics: ACL-IJCNLP 2021},
  pages={1489--1503},
  year={2021},
  doi={10.18653/v1/2021.findings-acl.130}
}

@inproceedings{oncoco2026,
  title={{OnCoCo 1.0}: A Public Dataset for Fine-Grained Message Classification in Online Counseling Conversations},
  author={Albrecht, Jens and Lehmann, Robert and Poltermann, Aleksandra and Rudolph, Eric and Steigerwald, Philipp and Stieler, Mara},
  booktitle={Proceedings of the Workshop on Social Context and Integrating NLP and Psychology to Study Social Interactions (SoCon-NLPSI), co-located with LREC 2026},
  year={2026},
  address={Palma de Mallorca, Spain},
  note={To appear}
}

@book{engelhardt2021,
  title={Lehrbuch {O}nlineberatung},
  author={Engelhardt, Emily M.},
  year={2021},
  edition={2},
  publisher={Vandenhoeck \& Ruprecht},
  address={G{\"o}ttingen}
}

@inproceedings{frohmann2024sat,
  title={Segment Any Text: A Universal Approach for Robust, Efficient and Adaptable Sentence Segmentation},
  author={Frohmann, Markus and Sterner, Igor and Vuli{\'c}, Ivan and Minixhofer, Benjamin and Schedl, Markus},
  booktitle={Proceedings of the 2024 Conference on Empirical Methods in Natural Language Processing (EMNLP)},
  pages={11908--11941},
  year={2024},
  address={Miami, Florida},
  doi={10.18653/v1/2024.emnlp-main.665}
}

@book{cohen1988statistical,
  title={Statistical Power Analysis for the Behavioral Sciences},
  author={Cohen, Jacob},
  year={1988},
  edition={2},
  publisher={Lawrence Erlbaum Associates},
  address={Hillsdale, NJ}
}

@article{niederhoffer2002linguistic,
  title={Linguistic Style Matching in Social Interaction},
  author={Niederhoffer, Kate G. and Pennebaker, James W.},
  journal={Journal of Language and Social Psychology},
  volume={21},
  number={4},
  pages={337--360},
  year={2002},
  publisher={Sage Publications}
}

@article{massey1951kolmogorov,
  title={The Kolmogorov-Smirnov test for goodness of fit},
  author={Massey Jr., Frank J.},
  journal={Journal of the American Statistical Association},
  volume={46},
  number={253},
  pages={68--78},
  year={1951},
  publisher={Taylor \& Francis}
}

@inproceedings{zheng2023judging,
  title={Judging {LLM}-as-a-Judge with {MT}-Bench and Chatbot Arena},
  author={Zheng, Lianmin and Chiang, Wei-Lin and Sheng, Ying and Zhuang, Siyuan and Wu, Zhanghao and Zhuang, Yonghao and Lin, Zi and Li, Zhuohan and Li, Dacheng and Xing, Eric P. and Zhang, Hao and Gonzalez, Joseph E. and Stoica, Ion},
  booktitle={Advances in Neural Information Processing Systems 36 (NeurIPS), Datasets and Benchmarks Track},
  year={2023}
}

@inproceedings{pillutla2021mauve,
  title={{MAUVE}: Measuring the Gap Between Neural Text and Human Text using Divergence Frontiers},
  author={Pillutla, Krishna and Swayamdipta, Swabha and Zellers, Rowan and Thickstun, John and Welleck, Sean and Choi, Yejin and Harchaoui, Zaid},
  booktitle={Advances in Neural Information Processing Systems 34 (NeurIPS)},
  year={2021}
}

@article{spearman1910,
  author  = {Spearman, Charles},
  title   = {Correlation calculated from faulty data},
  journal = {British Journal of Psychology},
  volume  = {3},
  number  = {3},
  pages   = {271--295},
  year    = {1910}
}

@article{brown1910,
  author  = {Brown, William},
  title   = {Some experimental results in the correlation of mental abilities},
  journal = {British Journal of Psychology},
  volume  = {3},
  number  = {3},
  pages   = {296--322},
  year    = {1910}
}

@article{esteban2017realvalued,
  author  = {Esteban, Crist{\'o}bal and Hyland, Stephanie L. and R{\"a}tsch, Gunnar},
  title   = {Real-valued (medical) time series generation with recurrent conditional {GAN}s},
  journal = {arXiv preprint arXiv:1706.02633},
  year    = {2017}
}

@article{bradley1952rank,
  author  = {Bradley, Ralph Allan and Terry, Milton E.},
  title   = {Rank analysis of incomplete block designs: {I}. {The} method of paired comparisons},
  journal = {Biometrika},
  volume  = {39},
  number  = {3/4},
  pages   = {324--345},
  year    = {1952}
}

@misc{mistral2024ministral,
  author       = {{Mistral AI}},
  title        = {Un {Ministral}, des {Ministraux}: Introducing the world's best edge models},
  year         = {2024},
  howpublished = {\url{https://mistral.ai/news/ministraux}}
}

@misc{mistral2025small,
  author       = {{Mistral AI}},
  title        = {Mistral {Small} 3.2 (24{B} {Instruct} 2506)},
  year         = {2025},
  howpublished = {\url{https://huggingface.co/mistralai/Mistral-Small-3.2-24B-Instruct-2506}}
}

@article{benjamini1995controlling,
  author    = {Benjamini, Yoav and Hochberg, Yosef},
  title     = {Controlling the false discovery rate: a practical and powerful approach to multiple testing},
  journal   = {Journal of the Royal Statistical Society: Series B (Methodological)},
  volume    = {57},
  number    = {1},
  pages     = {289--300},
  year      = {1995},
  publisher = {Wiley}
}

\appendix

\setlength{\textfloatsep}{6pt plus 2pt minus 2pt}
\setlength{\intextsep}{5pt plus 2pt minus 2pt}
\setlength{\floatsep}{6pt plus 2pt minus 2pt}
\setlength{\abovecaptionskip}{3pt}
\setlength{\belowcaptionskip}{2pt}
\renewcommand{\topfraction}{0.95}
\renewcommand{\bottomfraction}{0.9}
\renewcommand{\textfraction}{0.06}
\renewcommand{\floatpagefraction}{0.85}
\renewcommand{\dbltopfraction}{0.95}
\renewcommand{\dblfloatpagefraction}{0.8}


\section{Statistical Methodology}
\label{app:stat_method}

Two measures compare how counsellor strategies and client emotions are distributed across the proxy and the real data. The Jensen--Shannon divergence (JSD) quantifies how different two distributions are and Cram\'er's~$V$ gives a standardised effect size.
Let $P_P$ and $P_R$ denote the label distributions of the proxy and real corpus over $k$ categories, with $P_{P,i}$ and $P_{R,i}$ the proportions of category~$i$.
The reported proxy--real value $\mathrm{JSD}^{\mathrm{RP}}$ is symmetric in its arguments. Within-real split-half values are $\mathrm{JSD}^{\mathrm{RR}}$ (Appendix~\ref{app:splithalf}).

\subsection{Jensen--Shannon Divergence}

JSD is the primary distributional distance for three reasons.
It is \textbf{symmetric}, so proxy--real and real--proxy agree and the within-real split-half null is well defined.
It is \textbf{bounded} in $[0,\log_2 2] = [0,1]$, so values compare directly across taxonomies of very different cardinalities ($k=7$ to $k=66$).
And it stays \textbf{finite when one distribution puts zero mass on a category the other observes}, a routine situation in small corpora that makes KL divergence infinite.
JSD builds on the Kullback--Leibler divergence,
\begin{equation}
D_{\mathrm{KL}}(P_P \| P_R) \;=\; \sum_{i=1}^{k} P_{P,i} \, \log_2 \frac{P_{P,i}}{P_{R,i}}.
\label{eq:kl}
\end{equation}
$D_{\mathrm{KL}}$ is asymmetric. JSD symmetrises it through a mixture $M = \tfrac{1}{2}(P_P + P_R)$,
\begin{equation}
\begin{split}
\mathrm{JSD}(P_P \| P_R) &= \tfrac{1}{2}\,D_{\mathrm{KL}}(P_P \| M)\\
&\quad + \tfrac{1}{2}\,D_{\mathrm{KL}}(P_R \| M).
\end{split}
\label{eq:jsd}
\end{equation}
With $\log_2$, JSD equals zero when the distributions are identical and one when their supports are disjoint. 95\,\% CIs are obtained by resampling conversations with replacement (10{,}000 iterations), accounting for within-conversation correlation (intraclass correlation $\approx .03$--$.04$ by taxonomy).

\subsection{Cram\'er's $V$}

Cram\'er's~$V$ expresses the distributional difference as a standardised effect size,
\begin{equation}
V \;=\; \sqrt{\frac{\chi^2}{n \cdot \min(r-1,\, k-1)}},
\label{eq:cramer}
\end{equation}
where $\chi^2$ is Pearson's statistic, $n$ the total observations, $r$ the number of groups and $k$ the number of categories.
With $r = 2$ corpora this simplifies to $V = \sqrt{\chi^2 / n}$, bounded in $[0,1]$ ($0$ for identical category distributions, $1$ when corpus membership is fully determined).
\citet{cohen1988statistical} sets small/medium/large for $w$ at ${.1}/{.3}/{.5}$ and cautions that a constant $w$ reads differently as the table grows (\S7.2.3; his conversion is $\phi' = w/\sqrt{r-1}$, formula 7.2.6).
Here the thresholds are divided by $\sqrt{k-1}$ instead, a deliberately stricter convention, giving $\approx .04/{.11}/{.18}$ at $k\!=\!9$ and $\approx .01/{.04}/{.06}$ at $k\!=\!66$.
$V$ is reported raw and judged against these stricter thresholds, not the $df\!=\!1$ benchmark.

\subsection{Design-Effect Adjusted CIs}

Spans within a conversation are correlated, so per-category CIs use a cluster-adjusted standard error,
\begin{align}
\hat{\sigma} &\;=\; \sqrt{\frac{p\,(1-p)}{n} \cdot \mathrm{DEFF}}, \label{eq:sigma}\\[2pt]
\mathrm{DEFF} &\;=\; 1 + (\bar{n}_{\mathrm{conv}} - 1) \cdot \rho, \label{eq:deff_app}
\end{align}
where $p$ is the category proportion, $n$ the total span count, $\bar{n}_{\mathrm{conv}}$ the mean spans per conversation and $\rho$ the within-conversation intraclass correlation, measured per taxonomy ($.033$ OnCoCo, $.040$ Ekman, $.031$ GoEmotions).
Per-category tables report $\pm\,1.96\hat{\sigma}$ (95\,\% CI half-width).
Overlapping intervals suggest the difference may be within sampling uncertainty.

\subsection{Real Split-Half Reference Distribution}
\label{app:splithalf}

To characterise the within-real noise band, the real data’s 124 conversations are partitioned into two equally sized halves $R_1^{(b)}, R_2^{(b)}$ at $B = 1{,}000$ random split points.
For each split $b$,
\begin{equation}
\mathrm{JSD}^{\mathrm{RR}}_{(b)} \;=\; \mathrm{JSD}\bigl(P_{R_1^{(b)}} \;\big\|\; P_{R_2^{(b)}}\bigr),
\label{eq:split_jsd}
\end{equation}
yielding the empirical distribution $\{\mathrm{JSD}^{\mathrm{RR}}_{(b)}\}_{b=1}^{B}$ with mean $\mathrm{JSD}^{\mathrm{RR}}$.
Splitting at the conversation level rather than the span level preserves within-conversation correlation, so the distribution is the real data’s own noise band.
The proxy--real divergence $\mathrm{JSD}^{\mathrm{RP}}$ is placed on it by its empirical percentile (Table~\ref{tab:splithalf}), a value below the 95th percentile sitting inside the noise band.
The distribution is right-skewed and bounded below at zero, so the percentile is reported rather than a parametric $z$.

The conversation-progress bins (\S\ref{sec:bins}) and transitions (\S\ref{sec:dyad}) use the 48 complete real threads, where position and order are well defined. Including the early drop-offs changes little: the corpus pooled proxy--real JSD is $\shRPCo$ (counsellor) and $\shRPEk$ (emotion) over all 124, against $.0037$ and $.0026$ over the 48 and the two real distributions differ by only $\mathrm{JSD} = .0004$ on both lenses.
Per bin the drop-offs leave a small trace in the \textit{Formalities}-heavy edge bins of the counsellor lens.
A short thread is largely its opening and closing.
Against all 124 the pooled strategy gap rises in the opening bin from $.0030$ to $.0082$ and in the closing from $.0101$ to $.0207$, while the inner three bins and every emotion bin move by at most $.005$.
The complete-thread reference of \S\ref{sec:bins} gives the cleaner per-bin picture.

A second variant removes a residual sample-size confound. The 62/62 split-half has a wider noise band than the actual proxy--real comparison ($n_P\!=\!86$, $n_R\!=\!124$), so a further bootstrap null draws both groups with replacement from the 124 real conversations at the matched sizes ($B\!=\!1{,}000$). The resulting band is tighter and confirms the ordering in Table~\ref{tab:splithalf}.
Under this $n$-matched null every proxy--real value exceeds the 95th percentile, at the 9-category resolution $\mathrm{JSD}^{\mathrm{RP}}\!=\!\shRPCo$ vs.\ $\NmcoPNF$, finer taxonomies progressively further out (Table~\ref{tab:splithalf_nm}).
All nonetheless remain several-fold to an order of magnitude below the cross-domain references on the shared native basis (Appendix~\ref{app:baselines}).

\begin{table}[h]
\centering
\scriptsize
\begin{tabular*}{\columnwidth}{@{\extracolsep{\fill}}lccc@{}}
\toprule
\textbf{Taxonomy} & $\mathrm{JSD}^{\mathrm{RP}}$ & $\mathrm{JSD}^{\mathrm{RR\text{-}nm}}$ & $\mathrm{P95}^{\mathrm{RR\text{-}nm}}$ \\
\midrule
OnCoCo Co.\ (9) & \shRPCo & \NmcoMean & \NmcoPNF \\
Ekman (7)       & \shRPEk & \NmEkMean & \NmEkPNF \\
GoEmotions (28) & \shRPGo & \NmLalkMean & \NmLalkPNF \\
OnCoCo all (66) & \shRPAll & \NmAllMean & \NmAllPNF \\
\bottomrule
\end{tabular*}
\caption{$n$-matched null (both groups resampled from the real data, $n_P\!=\!86$, $n_R\!=\!124$); every proxy--real JSD exceeds its 95th percentile $\mathrm{P95}^{\mathrm{RR\text{-}nm}}$.}
\label{tab:splithalf_nm}
\end{table}

\begin{table}[h]
\centering
\scriptsize
\begin{tabular*}{\columnwidth}{@{\extracolsep{\fill}}lccc@{}}
\toprule
\textbf{Taxonomy} & $\mathrm{JSD}^{\mathrm{RP}}$ & $\mathrm{JSD}^{\mathrm{RR}}$ & \textbf{Pctl.} \\
\midrule
OnCoCo Co.\ (9)   & \shRPCo & \shRRCo & \shPctlCo \\
OnCoCo Co.\ (38)  & \shRPCoLeaf & \shRRCoLeaf & \shPctlCoLeaf \\
OnCoCo all (66)   & \shRPAll & \shRRAll & \shPctlAll \\
Ekman (7)         & \shRPEk & \shRREk & \shPctlEk \\
GoEmotions (28)   & \shRPGo & \shRRGo & \shPctlGo \\
\bottomrule
\end{tabular*}
\caption{Proxy--real JSD vs.\ the within-real split-half (1{,}000 splits); Pctl.\ = its percentile.
Bracketed sizes are nominal, the real data observes 32 of 38 and 59 of 66.}
\label{tab:splithalf}
\end{table}

\subsection{Multiple-Comparison Control}
\label{app:fdr}

The within-message analysis (\S\ref{sec:dyad}) tests every transition cell at once, up to $9\times8$ off-diagonal counsellor pairs and $7\times6$ client pairs.
With that many tests at the 95\,\% level, some cells cross their band by chance.
The \emph{false discovery rate} (FDR) is the expected share of falsely flagged cells among all flagged ones and Benjamini--Hochberg control \cite{benjamini1995controlling} caps it at $q = \fdrQ$.
Each eligible cell (off-diagonal, with at least $10$ combined transitions) carries a two-sided empirical $p$-value, the fraction of the $2{,}000$ real-data resamples whose $|\Delta|$ reaches the observed $|\Delta|$.
Ordering these as $p_{(1)} \leq \dots \leq p_{(m)}$, the procedure flags every cell up to the largest rank $k$ for which
\begin{equation}
p_{(k)} \;\leq\; \frac{k}{m}\,q.
\label{eq:bh}
\end{equation}
At most a fraction $q$ of the flagged shifts are expected to be false, so a surviving cell is a shift that resampling the real data rarely produces.
The level $q = \fdrQ$ is a convention, not derived from the data.
FDR analyses use $0.05$ or $0.10$ and $0.10$ is the more lenient choice, common in exploratory multiple testing.
Even at this lenient level few cells survive, as most differences sit inside the noise band.


\section{JSD Baselines}
\label{app:baselines}

The proxy--real divergence is compared against two reference points: a split-half baseline (expected noise within the real data itself) and cross-domain counselling corpora (ESConv \cite{liu2021ESConv} and AnnoMI \cite{wu2022annomi}), all annotated with the same pipeline.
ESConv and AnnoMI are counselling, so the contrast stays fair.
They differ in language, modality and format, so they mark what a genuinely different corpus looks like.
An in-domain German e-mail corpus would sit almost on the real data and give the scale no upper end.
Field-foreign text such as news would be trivially far.

English enters only through these two reference corpora; the proxy--real validation itself is German throughout.
The OnCoCo classifier handles the English input in-distribution, having been trained on bilingual German--English counselling data (\S\ref{sec:annot_oncoco}).
The emotion classifier is a multilingual XLM-RoBERTa fine-tuned on German only and labels the English corpora in zero-shot cross-lingual transfer.
On the original English GoEmotions test set it reaches $F_1 = .38$ over the 28 categories (vs.\ $.45$ on German) and $.49$ after the Ekman collapse.
The emotion reference values are therefore noisier than the German measurements but remain meaningful.
The anchors are read as the distance to a genuinely different corpus, in which language, modality and counselling format differ together, not as a pure domain distance.
The tables in this appendix use the native, coarser OnCoCo message segmentation rather than the finer SaT-enriched spans of the main text, so absolute JSDs are larger here (e.g.\ $.020$ vs.\ $\shRPCo$ for counsellor strategies, $.017$/$.049$ vs.\ $\shRPEk$/$\shRPGo$ for Ekman/GoEmotions).
The position of proxy--real relative to the noise floor and the cross-domain references is unchanged.

Table~\ref{tab:jsd_oncoco} reports counsellor strategy divergence at four OnCoCo granularity levels.
After block-merging, the proxy--real JSDs sit at one to two-and-a-half times the within-real noise floor and roughly an order of magnitude below the AnnoMI/ESConv references, with proxy--real $V \leq .25$ throughout while cross-domain $V$ jumps to $.38$--$.67$.

\begin{table}[h]
\centering
\resizebox{\columnwidth}{!}{%
\scriptsize
\begin{tabular}{@{}l cc cc cc cc@{}}
\toprule
& \multicolumn{2}{c}{\textbf{Co.\ (9)}} & \multicolumn{2}{c}{\textbf{Co.\ (38)}} & \multicolumn{2}{c}{\textbf{Cl.\ OnCoCo (28)}} & \multicolumn{2}{c}{\textbf{all (66)}} \\
\cmidrule(lr){2-3} \cmidrule(lr){4-5} \cmidrule(lr){6-7} \cmidrule(lr){8-9}
\textbf{Comparison} & JSD & $V$ & JSD & $V$ & JSD & $V$ & JSD & $V$ \\
\midrule
Proxy--Real      & .020 & .16 & .044 & .23 & .039 & .21 & .045 & .23 \\
Real split-half   & .009 & .10 & .035 & .20 & .037 & .21 & .036 & .20 \\
Proxy--AnnoMI    & .215 & .52 & .415 & .67 & .363 & .63 & .389 & .65 \\
Proxy--ESConv    & .213 & .43 & .400 & .48 & .257 & .39 & .327 & .44 \\
Real--AnnoMI      & .270 & .57 & .445 & .66 & .443 & .67 & .445 & .67 \\
Real--ESConv      & .307 & .47 & .466 & .50 & .322 & .38 & .399 & .45 \\
\bottomrule
\end{tabular}%
}
\caption{OnCoCo divergence at four granularities (native annotations).
Co.\ = counsellor, Cl.\ = client, all = both speakers; $V$ = Cram\'er's $V$ (Appendix~\ref{app:stat_method}).}
\label{tab:jsd_oncoco}
\end{table}

Table~\ref{tab:jsd_emo} shows the same pattern for client emotions.
At GoEmotions resolution the ordering is proxy--real $\ll$ proxy--ESConv $<$ proxy--AnnoMI and the proxy--real gap ($.049$) sits close to the real data’s own native noise floor ($.041$).

\begin{table}[h]
\centering
\scriptsize
\begin{tabular*}{\columnwidth}{@{\extracolsep{\fill}}l cc cc@{}}
\toprule
& \multicolumn{2}{c}{\textbf{Ekman (7)}} & \multicolumn{2}{c}{\textbf{GoEmotions (28)}} \\
\cmidrule(lr){2-3} \cmidrule(lr){4-5}
\textbf{Comparison} & JSD & $V$ & JSD & $V$ \\
\midrule
Proxy--Real      & .017 & .14 & .049 & .24 \\
Real split-half   & .008 & .10 & .041 & .22 \\
Proxy--AnnoMI    & .017 & .12 & .199 & .46 \\
Proxy--ESConv    & .022 & .07 & .126 & .20 \\
Real--AnnoMI      & .051 & .17 & .290 & .50 \\
Real--ESConv      & .044 & .07 & .172 & .16 \\
\bottomrule
\end{tabular*}
\caption{Client emotion divergence (native annotations); proxy--real $\ll$ cross-domain at GoEmotions-28, washing out at Ekman-7.}
\label{tab:jsd_emo}
\end{table}

Table~\ref{tab:jsd_bins} breaks the comparison down by conversation-progress quintile.
All per-bin $V \leq .21$, confirming that the similarity holds at every stage of the conversation.

\begin{table}[h]
\centering
\resizebox{\columnwidth}{!}{%
\scriptsize
\begin{tabular}{@{}l cc cc cc cc@{}}
\toprule
& \multicolumn{2}{c}{\textbf{Co.\ (9)}} & \multicolumn{2}{c}{\textbf{Co.\ (38)}} & \multicolumn{2}{c}{\textbf{Ekman}} & \multicolumn{2}{c}{\textbf{GoEmotions}} \\
\cmidrule(lr){2-3} \cmidrule(lr){4-5} \cmidrule(lr){6-7} \cmidrule(lr){8-9}
\textbf{Bin} & JSD & $V$ & JSD & $V$ & JSD & $V$ & JSD & $V$ \\
\midrule
\perbinRowsApp
\bottomrule
\end{tabular}%
}
\caption{Per-bin proxy--real divergence over five quintiles (block-merged); all $V \leq .21$, last row = corpus level.}
\label{tab:jsd_bins}
\end{table}


\section{MAUVE Setup}
\label{app:mauve}

Both MAUVE analyses (the generative check in Section~\ref{sec:downstream} and the corpus-level raw-text companion below) use the reference implementation \cite{pillutla2021mauve}.
MAUVE is bounded in $[0,1]$, $1$ = the two text distributions are indistinguishable, $0$ = they share no common region.
The score is the area under the KL-divergence curve between the two sides' cluster histograms, obtained by quantising the pooled embeddings.
Each GEMCo message or candidate reply is one sample; the cross-domain corpora contribute one utterance per sample (600 sampled from each).
Features are the last-token hidden state of \texttt{Ministral-8B-Instruct-2410} (fp16, texts truncated at 512 tokens).
The model is multilingual, so the English reference corpora share the embedding space.
Every run uses identical settings (25 quantisation buckets, fixed seed), so the values are directly comparable.

\paragraph{Generative check (\S\ref{sec:downstream}).} At each of the \dsNturns{} judged counsellor turns there are three replies: the human counsellor's, the proxy-tuned model's and the off-the-shelf model's.
The 124 real conversations split into two random halves of 62, about 162 turns per side.
Each half contributes its conversations' turns, so disjoint conversation sets sit on the two sides and no model reply meets the human reply of its own turn.
The human replies of half one are compared with three second halves: the human replies (ceiling), the proxy-tuned replies and the off-the-shelf replies at those turns.
Embeddings, quantisation (25 buckets) and seed are shared across the rows, which differ only in whose replies stand in the second half.
Table~\ref{tab:genval_mauve} reports mean $\pm$ standard deviation over 50 splits.

At full sample size (325 vs.\ 325, not size-matched) the values are $\mauveHumanSft$ (human--Proxy-SFT), $\mauveHumanInstruct$ (human--instruct) and $\mauveSftInstruct$ (Proxy-SFT--instruct).
MAUVE rises with sample size, so these are not comparable to the split-half ceiling.

\paragraph{Corpus-level raw-text check.}

At the whole-message level the within-real split-half reference is $\mauveRR \pm \mauveRRsd$ (minimum $\mauveRRmin$).
Against the real data, pooled proxy reaches $\mauveRP$, GEMCo-A $\mauveA$ and GEMCo-B $\mauveB$.
The cross-domain anchors sit at $\mauvePEsconv$--$\mauveREsconv$ (proxy/real vs.\ ESConv) and $\mauvePAnnomi$--$\mauveRAnnomi$ (vs.\ AnnoMI).
At the surface level proxy is clearly distinguishable from the real data, yet two orders of magnitude closer to it than the cross-domain corpora.
The label-based validation of Sections~\ref{sec:corpus_validation}--\ref{sec:per_category} measures what remains once this surface register drops out.

Two controls separate length from register.
Splitting the real data at its median message length gives $\mathrm{MAUVE} = \mauveRealShortLong$ between the short and the long half, so the embedding is strongly length-sensitive.
Matching the real data to GEMCo-B's length distribution (decile matching, drawn with replacement) nonetheless leaves the GEMCo-B--real value at $\mauveBLenmatch$ (vs.\ $\mauveB$ unmatched).
GEMCo-B's distance therefore reflects its role-played register and the deliberately shared case vignettes (\S\ref{sec:corpus}), not message length.


\section{Per-Subcorpus Distributions With Confidence Intervals}
\label{app:full_dists}

Table~\ref{tab:cat_dist} gives the per-subcorpus distributions in compact form, for GEMCo-A, GEMCo-B and the real data.
Tables~\ref{tab:per_cat_full} and~\ref{tab:per_cat_full_go} add 95\,\% design-effect-adjusted confidence intervals (\textpm 1.96$\hat{\sigma}$, $\rho = .033/.040/.031$ by taxonomy; Eq.~\ref{eq:sigma}) for every category at all three resolutions: counsellor strategies (9~OnCoCo categories), Ekman emotions (7) and GoEmotions (28).
Table~\ref{tab:per_cat} in the main text picks the three rows where GEMCo-B's CI lies fully outside the real data’s.
Every other category overlaps within its interval across GEMCo-A, GEMCo-B and the real data.

\begin{table}[h]
\centering
\footnotesize
\setlength{\tabcolsep}{3pt}
\begin{tabular*}{\columnwidth}{@{\extracolsep{\fill}}lrrr@{}}
\toprule
\textbf{Category} & \textbf{GEMCo-A} & \textbf{GEMCo-B} & \textbf{Real} \\
\midrule
\multicolumn{4}{@{}l}{\textit{Counsellor strategy (OnCoCo 9)}} \\
\quad Formalities (beginning) & 6.5 & 8.0 & 7.5 \\
\quad Formalities (conclusion) & 8.9 & 8.5 & 8.9 \\
\quad Moderation & 13.7 & \textbf{11.0} & 16.1 \\
\quad Other & 9.4 & 8.1 & 9.2 \\
\quad Analysis \& Clarification & 18.6 & 24.7 & 21.3 \\
\quad Agreement on Objectives & 6.2 & 7.3 & 5.8 \\
\quad Creating Motivation & 10.6 & 10.1 & 8.0 \\
\quad Resource Activation & 6.9 & 4.7 & 4.7 \\
\quad Help \& Problem Solving & 19.3 & 17.5 & 18.7 \\
\midrule
\multicolumn{4}{@{}l}{\textit{Client emotion (Ekman 7)}} \\
\quad joy & 29.5 & 25.4 & 24.5 \\
\quad sadness & 11.0 & \textbf{6.9} & 12.7 \\
\quad fear & 3.6 & 3.9 & 4.2 \\
\quad anger & 7.3 & 8.2 & 8.7 \\
\quad surprise & 12.9 & \textbf{17.1} & 12.1 \\
\quad disgust & 0.6 & 0.6 & 0.8 \\
\quad neutral & 35.2 & 37.8 & 36.9 \\
\bottomrule
\end{tabular*}

\caption{Label distribution (\%) by subcorpus: counsellor strategy (OnCoCo 9) over counsellor blocks and client emotion (Ekman 7) over client blocks.
Proxy pools GEMCo-A and GEMCo-B; bold = 95\,\% CI fully outside the real data’s.}
\label{tab:cat_dist}
\end{table}

\begin{table}[h]
\centering
\scriptsize
\setlength{\tabcolsep}{2.5pt}
\begin{tabular*}{\linewidth}{@{\extracolsep{\fill}}lrrr@{}}
\toprule
\textbf{Category} & \textbf{GEMCo-A} & \textbf{GEMCo-B} & \textbf{Real} \\
\midrule
\multicolumn{4}{@{}l}{\textit{Counsellor strategies (OnCoCo 9)}} \\
\quad Formalities (beginning) & 6.5\,\textpm\,1.4 & 8.0\,\textpm\,2.0 & 7.5\,\textpm\,1.1 \\
\quad Formalities (conclusion) & 8.9\,\textpm\,1.7 & 8.5\,\textpm\,2.1 & 8.9\,\textpm\,1.2 \\
\quad Moderation & 13.7\,\textpm\,2.0 & 11.0\,\textpm\,2.3 & 16.1\,\textpm\,1.5 \\
\quad Other & 9.4\,\textpm\,1.7 & 8.1\,\textpm\,2.0 & 9.2\,\textpm\,1.2 \\
\quad Analysis \& Clarification & 18.6\,\textpm\,2.3 & 24.7\,\textpm\,3.2 & 21.3\,\textpm\,1.7 \\
\quad Agreement on Objectives & 6.2\,\textpm\,1.4 & 7.3\,\textpm\,1.9 & 5.8\,\textpm\,1.0 \\
\quad Creating Motivation & 10.6\,\textpm\,1.8 & 10.1\,\textpm\,2.2 & 8.0\,\textpm\,1.1 \\
\quad Resource Activation & 6.9\,\textpm\,1.5 & 4.7\,\textpm\,1.6 & 4.7\,\textpm\,0.9 \\
\quad Help \& Problem Solving & 19.3\,\textpm\,2.3 & 17.5\,\textpm\,2.8 & 18.7\,\textpm\,1.6 \\
\midrule
\multicolumn{4}{@{}l}{\textit{Client emotions (Ekman 7)}} \\
\quad joy & 29.5\,\textpm\,3.1 & 25.4\,\textpm\,3.7 & 24.5\,\textpm\,2.0 \\
\quad sadness & 11.0\,\textpm\,2.1 & 6.9\,\textpm\,2.2 & 12.7\,\textpm\,1.6 \\
\quad fear & 3.6\,\textpm\,1.3 & 3.9\,\textpm\,1.6 & 4.2\,\textpm\,0.9 \\
\quad anger & 7.3\,\textpm\,1.7 & 8.2\,\textpm\,2.3 & 8.7\,\textpm\,1.3 \\
\quad surprise & 12.9\,\textpm\,2.3 & 17.1\,\textpm\,3.2 & 12.1\,\textpm\,1.5 \\
\quad disgust & 0.6\,\textpm\,0.5 & 0.6\,\textpm\,0.7 & 0.8\,\textpm\,0.4 \\
\quad neutral & 35.2\,\textpm\,3.2 & 37.8\,\textpm\,4.1 & 36.9\,\textpm\,2.3 \\
\bottomrule
\end{tabular*}
\caption{Full per-subcorpus percentages (\,\textpm 1.96$\hat{\sigma}$): counsellor strategy (OnCoCo 9) and client Ekman emotion (7).}
\label{tab:per_cat_full}
\end{table}

\begin{table}[h]
\centering
\scriptsize
\setlength{\tabcolsep}{2.5pt}
\begin{tabular*}{\linewidth}{@{\extracolsep{\fill}}lrrr@{}}
\toprule
\textbf{Category} & \textbf{GEMCo-A} & \textbf{GEMCo-B} & \textbf{Real} \\
\midrule
\multicolumn{4}{@{}l}{\textit{Client emotions (GoEmotions 28)}} \\
\quad admiration & 3.6\,\textpm\,1.1 & 2.8\,\textpm\,1.3 & 2.8\,\textpm\,0.7 \\
\quad amusement & 0.8\,\textpm\,0.5 & 0.7\,\textpm\,0.7 & 0.2\,\textpm\,0.2 \\
\quad approval & 6.3\,\textpm\,1.5 & 6.3\,\textpm\,1.9 & 6.4\,\textpm\,1.0 \\
\quad caring & 2.7\,\textpm\,1.0 & 2.2\,\textpm\,1.1 & 2.8\,\textpm\,0.7 \\
\quad desire & 2.2\,\textpm\,0.9 & 1.9\,\textpm\,1.1 & 1.7\,\textpm\,0.6 \\
\quad excitement & 1.1\,\textpm\,0.6 & 0.5\,\textpm\,0.6 & 0.7\,\textpm\,0.4 \\
\quad gratitude & 8.3\,\textpm\,1.7 & 7.9\,\textpm\,2.1 & 7.6\,\textpm\,1.1 \\
\quad joy & 3.1\,\textpm\,1.0 & 1.8\,\textpm\,1.0 & 2.3\,\textpm\,0.6 \\
\quad love & 5.1\,\textpm\,1.3 & 2.6\,\textpm\,1.2 & 3.0\,\textpm\,0.7 \\
\quad optimism & 3.1\,\textpm\,1.0 & 2.3\,\textpm\,1.2 & 2.2\,\textpm\,0.6 \\
\quad pride & 0.3\,\textpm\,0.3 & 0.3\,\textpm\,0.4 & 0.2\,\textpm\,0.2 \\
\quad relief & 0.5\,\textpm\,0.4 & 0.7\,\textpm\,0.7 & 0.5\,\textpm\,0.3 \\
\quad confusion & 4.0\,\textpm\,1.2 & 8.3\,\textpm\,2.1 & 3.7\,\textpm\,0.8 \\
\quad curiosity & 4.5\,\textpm\,1.3 & 4.0\,\textpm\,1.5 & 3.2\,\textpm\,0.7 \\
\quad realization & 3.8\,\textpm\,1.2 & 3.8\,\textpm\,1.5 & 4.5\,\textpm\,0.9 \\
\quad surprise & 0.5\,\textpm\,0.4 & 1.2\,\textpm\,0.8 & 0.6\,\textpm\,0.3 \\
\quad anger & 0.7\,\textpm\,0.5 & 0.4\,\textpm\,0.5 & 1.1\,\textpm\,0.4 \\
\quad annoyance & 3.1\,\textpm\,1.0 & 3.6\,\textpm\,1.4 & 3.2\,\textpm\,0.7 \\
\quad disappointment & 3.3\,\textpm\,1.1 & 2.4\,\textpm\,1.2 & 4.0\,\textpm\,0.8 \\
\quad disapproval & 2.9\,\textpm\,1.0 & 3.6\,\textpm\,1.4 & 3.9\,\textpm\,0.8 \\
\quad disgust & 0.5\,\textpm\,0.4 & 0.6\,\textpm\,0.6 & 0.7\,\textpm\,0.4 \\
\quad embarrassment & 0.4\,\textpm\,0.4 & 0.4\,\textpm\,0.5 & 0.6\,\textpm\,0.3 \\
\quad fear & 2.0\,\textpm\,0.8 & 2.2\,\textpm\,1.1 & 2.7\,\textpm\,0.7 \\
\quad grief & 0.0\,\textpm\,0.1 & 0.0\,\textpm\,0.0 & 0.1\,\textpm\,0.1 \\
\quad nervousness & 1.1\,\textpm\,0.6 & 1.4\,\textpm\,0.9 & 1.2\,\textpm\,0.5 \\
\quad remorse & 0.5\,\textpm\,0.4 & 0.2\,\textpm\,0.4 & 0.3\,\textpm\,0.2 \\
\quad sadness & 5.7\,\textpm\,1.4 & 3.3\,\textpm\,1.4 & 7.2\,\textpm\,1.1 \\
\quad neutral & 29.9\,\textpm\,2.8 & 34.5\,\textpm\,3.7 & 32.7\,\textpm\,2.0 \\
\bottomrule
\end{tabular*}
\caption{Full per-subcorpus percentages (\,\textpm 1.96$\hat{\sigma}$): client emotion at GoEmotions resolution (28 categories).}
\label{tab:per_cat_full_go}
\end{table}


\section{Client Emotion at GoEmotions Resolution}
\label{app:goemotions}

The main text reports client emotion at the seven-class Ekman level (\S\ref{sec:emotions}).
The finer 28-category resolution localises the residual proxy--real gap to specific affective categories.
Table~\ref{tab:ekman_map} gives the official grouping of the 28 GoEmotions categories onto Ekman's six basic emotions plus neutral.

\begin{table}[h]
\centering
\footnotesize
\begin{tabular*}{\columnwidth}{@{\extracolsep{\fill}}lp{0.7\columnwidth}@{}}
\toprule
\textbf{Ekman} & \textbf{GoEmotions categories} \\
\midrule
joy & admiration, amusement, approval, caring, desire, excitement, gratitude, joy, love, optimism, pride, relief \\
surprise & confusion, curiosity, realization, surprise \\
sadness & disappointment, embarrassment, grief, remorse, sadness \\
anger & anger, annoyance, disapproval \\
fear & fear, nervousness \\
disgust & disgust \\
neutral & neutral \\
\bottomrule
\end{tabular*}
\caption{Official GoEmotions$\to$Ekman grouping \cite{demszky2020goemotions} behind the seven-class client-emotion resolution (\S\ref{sec:annot_emotion}).}
\label{tab:ekman_map}
\end{table}

\begin{figure}[h]
    \centering
    \includegraphics[width=\columnwidth]{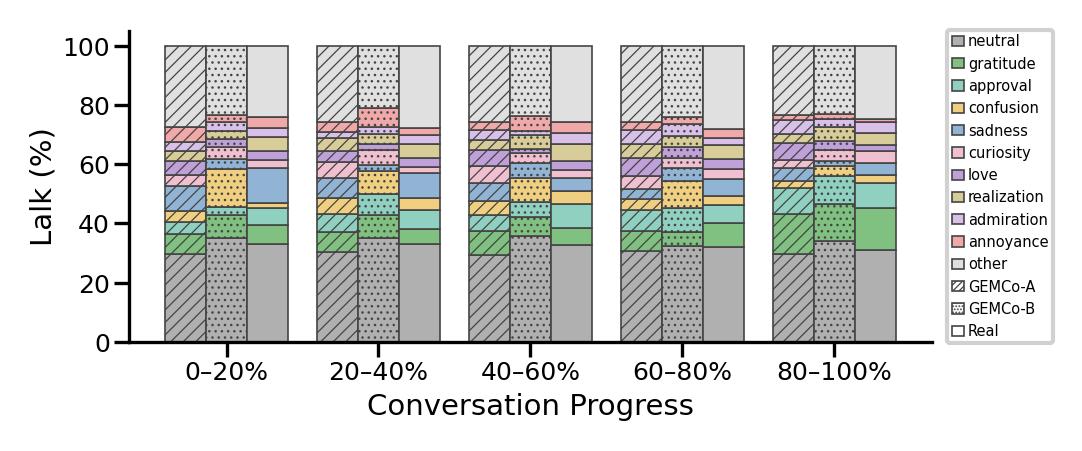}
    \caption{Client emotion (GoEmotions, 28 cat.) over conversation progress.
GEMCo-A hatched, GEMCo-B dotted, Real plain.}
    \label{fig:lalk_split}
\end{figure}

Figure~\ref{fig:lalk_split} shows the full GoEmotions distribution over conversation progress for GEMCo-A, GEMCo-B and the real data.
The corpus-level divergence sits at the \shPctlGo th percentile of the real data’s noise band ($\mathrm{JSD}^{\mathrm{RP}}_{\mathrm{GoEmotions}} = \shRPGo$, $V = \vGo$; Table~\ref{tab:splithalf}) and on the shared native basis stays several-fold below the cross-domain references ($.049$ vs.\ $.17$--$.29$; Table~\ref{tab:jsd_emo}).

Two production modes separate at this resolution.
GEMCo-A's \emph{narrative-closure} mode shows elevated \textit{love} ($+2.1$\,pp), which clears the real data’s CI although its coarser \textit{joy} form does not.
GEMCo-B's \emph{problem-fixation} mode shows elevated \textit{confusion} ($+4.6$\,pp) and reduced \textit{sadness} ($-3.9$\,pp), both outside the real data’s CI.
These map onto the \textit{joy} and \textit{surprise} shifts of Section~\ref{sec:per_category}.

\section{Counsellor Strategy at Leaf Resolution}
\label{app:counsellor_leaf}

\begin{figure}[h]
    \centering
    \includegraphics[width=\columnwidth]{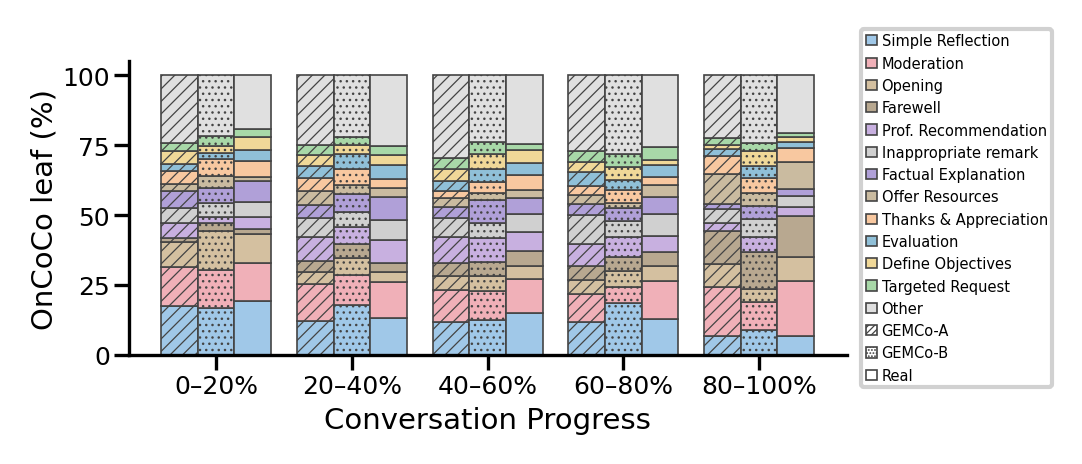}
    \caption{Counsellor strategy at OnCoCo leaf resolution (top 12 of 38 categories) over conversation progress.
GEMCo-A hatched, GEMCo-B dotted, Real plain.}
    \label{fig:co_leaf_split}
\end{figure}

The main text reports counsellor strategy at the nine top-level OnCoCo categories (\S\ref{sec:strategies}).
OnCoCo is hierarchical, each category subsuming finer acts down to 38 \emph{leaf} categories \cite{oncoco2026} and agreement holds at this finer resolution.
Figure~\ref{fig:co_leaf_split} shows the twelve most frequent leaf categories over conversation progress for GEMCo-A, GEMCo-B and Real.
Proxy and the real data track closely throughout and no leaf category becomes a subcorpus signature, unlike the client emotion modes of Appendix~\ref{app:goemotions}.

The divergence grows with resolution but remains small. At the 38-leaf level the gap rises to $\mathrm{JSD}^{\mathrm{RP}}_{\mathrm{OnCoCo}} = .044$ ($V = .23$, native annotations), an order of magnitude below the cross-domain references, whose JSD to ESConv and AnnoMI reaches $.40$--$.47$ ($V = .48$--$.67$; Table~\ref{tab:jsd_oncoco}).

\section{Within-Message Transitions}
\label{app:within_co}

\begin{figure}[h]
    \centering
    \begin{subfigure}[t]{\columnwidth}
        \centering
        \includegraphics[width=\columnwidth]{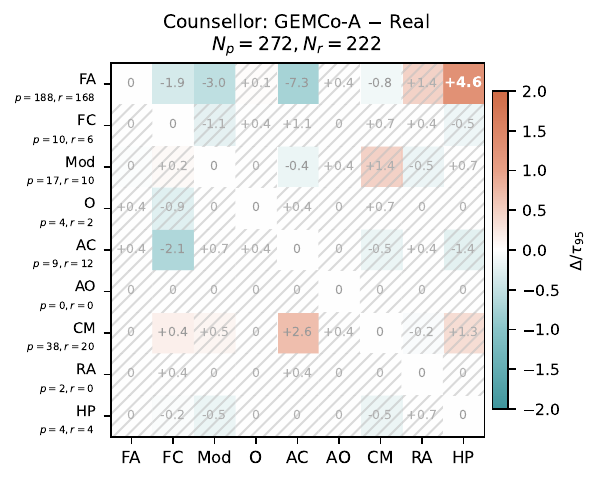}
        \caption{GEMCo-A counsellor vs.\ the real data ($\geq$5-message threads).}
        \label{fig:within_co_a}
    \end{subfigure}
    \vspace{2pt}
    \begin{subfigure}[t]{\columnwidth}
        \centering
        \includegraphics[width=\columnwidth]{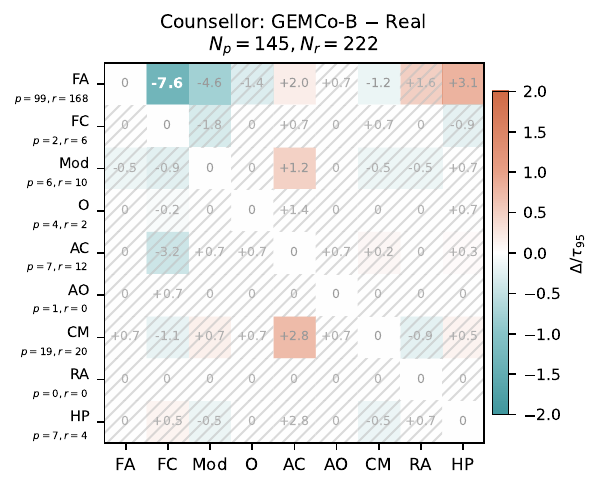}
        \caption{GEMCo-B counsellor vs.\ the real data ($\geq$5-message threads).}
        \label{fig:within_co_b}
    \end{subfigure}
    \caption{Counsellor strategy transitions vs.\ the 48 real threads with at least five messages (\S\ref{sec:dyad}).
Cell text = raw delta (pp), colour = $\Delta/\tau_{95}$, hatched = too sparse to test.}
    \label{fig:within_co}
\end{figure}

These matrices give the full per-cell results behind Section~\ref{sec:dyad}, where a cell is one ordered pair of consecutive blocks (row = current, column = next), the text is the raw proxy$-$real delta (pp) and the colour its standardised exceedance $\Delta/\tau_{95}$ of that cell's within-real noise band (hatched where too sparse to test).
The band $\tau_{95}$ is the 95th percentile of the proxy--real difference over $2{,}000$ real-data resamples.
Shifts past the band are flagged under Benjamini--Hochberg FDR control ($q = \fdrQ$, Appendix~\ref{app:fdr}).
Figure~\ref{fig:within_co} shows counsellor strategy transitions, where no cell is consistently shifted across the two subcorpora.
The client emotion transitions yield several FDR survivors: more movement around \textit{joy} in GEMCo-A and less traffic into and out of \textit{sadness} in GEMCo-B (\S\ref{sec:dyad}).

\section{OnCoCo Category Descriptions}
\label{app:categories}

Table~\ref{tab:cat_co} describes the nine OnCoCo top-level counsellor categories, with short corpus-style example utterances and English glosses. The four formal and five Impact-Factor categories are listed in \S\ref{sec:annot_oncoco}.

\begin{table*}[t]
\centering
\footnotesize
\renewcommand{\arraystretch}{1.1}
\begin{tabular}{@{}p{0.7cm}p{3.2cm}p{\dimexpr\textwidth-4.6cm\relax}@{}}
\toprule
\textbf{Code} & \textbf{Category} & \textbf{Description \& example} \\
\midrule
FA & Formalities (beginning) & Opening greeting, acknowledging the message.
\emph{``Liebe Maria, vielen Dank für deine Nachricht.''} [Dear Maria, thank you for your message.] \\
Mod & Moderation & Conversation management, structuring or signposting the reply.
\emph{``Ich gehe gleich auf deine einzelnen Fragen ein.''} [I'll address your individual questions in a moment.] \\
AC & Analysis \& Clarification & Probing or reflecting the client's situation back.
\emph{``Habe ich richtig verstanden, dass du dich von deinen Eltern unter Druck gesetzt fühlst?''} [Did I understand correctly that you feel pressured by your parents?] \\
AO & Agreement on Objectives & Proposing or agreeing on a counselling goal.
\emph{``Vielleicht konzentrieren wir uns zunächst auf das Gespräch mit deiner Mutter.''} [Perhaps we first focus on the conversation with your mother.] \\
CM & Creating Motivation & Encouragement, motivational reframing.
\emph{``Du hast schon einen wichtigen Schritt gemacht, indem du dich gemeldet hast.''} [You've already taken an important step by reaching out.] \\
RA & Resource Activation & Pointing the client to personal or social resources.
\emph{``Vielleicht könnte deine beste Freundin dir in dieser Situation Rückhalt geben.''} [Maybe your best friend could give you support in this situation.] \\
HP & Help \& Problem Solving & Concrete advice or suggested action.
\emph{``Du könntest einen Termin bei der Schulpsychologin vereinbaren.''} [You could make an appointment with the school psychologist.] \\
O  & Other & Statements outside the schema (meta-talk, asides).
\emph{``Ich kenne deine Stadt nicht persönlich, aber\,\dots''} [I don't know your city personally, but\,\dots] \\
FC & Formalities (conclusion) & Closing remarks, sign-off.
\emph{``Ich wünsche dir alles Gute.
Liebe Grüße, Anna.''} [I wish you all the best.
Warm regards, Anna.] \\
\bottomrule
\end{tabular}
\caption{OnCoCo top-level counsellor categories.}
\label{tab:cat_co}
\end{table*}


\section{Generative Validation Setup}
\label{app:genval}

\textbf{Fine-tuning.} The proxy-tuned model is \texttt{Mistral-Small-3.2-24B-Instruct-2506} with a QLoRA adapter (4-bit NF4, double quantisation, bf16 compute) trained on the 342 proxy counsellor turns (GEMCo-A + GEMCo-B) for 3 epochs: LoRA rank 64, $\alpha\!=\!128$, dropout 0.05 on all attention and MLP projections; learning rate $5\!\times\!10^{-5}$ (cosine, 3\% warmup); effective batch size 8 ($1\times8$ gradient accumulation); max sequence length 4096; \texttt{paged\_adamw\_32bit}; seed 42. The off-the-shelf baseline is the same base model without the adapter.

\textbf{Candidate generation.} All candidates use nucleus sampling at temperature 0.7, top-$p$ 0.95, up to 600 new tokens, seed 42 and no length control. Proxy-SFT keeps its training-time \texttt{User:}/\texttt{Counsellor:} template and system prompt. The off-the-shelf model uses the engineered counsellor system prompt below.

\begin{tcolorbox}[
  colback=clrPromptDE!7!white, colframe=clrPromptDE!75!black,
  title={\fontsize{8}{9.5}\selectfont\bfseries Proxy-SFT System Prompt --- German (verbatim) / English},
  fontupper=\fontsize{7.5}{9}\selectfont,
  breakable, boxrule=0.5pt,
  left=4pt, right=4pt, top=3pt, bottom=3pt, toptitle=2pt, bottomtitle=2pt,
]
Du bist eine erfahrene psychosoziale Beraterin in einer asynchronen E-Mail-Beratung.
Antworte mit der n\"achsten Beratenden-Nachricht an die ratsuchende Person --- im Stil, Ton und Umfang einer realen E-Mail-Beratungsantwort.
Keine Rollenangaben, keine Meta-Kommentare.
\tcbline
\textit{You are an experienced psychosocial counsellor in an asynchronous e-mail counselling service.
Reply with the next counsellor message to the help-seeking person --- in the style, tone and length of a real e-mail counselling reply.
No role labels, no meta comments.}
\end{tcolorbox}

\begin{tcolorbox}[
  colback=clrPromptDE!7!white, colframe=clrPromptDE!75!black,
  title={\fontsize{8}{9.5}\selectfont\bfseries Off-the-Shelf Counsellor System Prompt --- German (verbatim) / English},
  fontupper=\fontsize{7.5}{9}\selectfont,
  breakable, boxrule=0.5pt,
  left=4pt, right=4pt, top=3pt, bottom=3pt, toptitle=2pt, bottomtitle=2pt,
]
Du bist eine erfahrene psychosoziale Beraterin in einer asynchronen E-Mail-Beratung eines deutschen Online-Beratungsdienstes.
Du erh\"altst den bisherigen E-Mail-Verlauf zwischen einer ratsuchenden Person und der Beratung.
Schreibe ausschlie\ss lich die n\"achste Nachricht der Beratenden.\\[3pt]
\textbf{Orientiere dich an professioneller, systemisch--klientenzentrierter Beratungspraxis:}
\begin{itemize}[nosep, leftmargin=12pt]
  \item Greife konkret auf, was die ratsuchende Person zuletzt geschrieben hat; spiegle Gef\"uhle, ohne zu beschwichtigen oder zu \"ubertreiben.
  \item Dr\"ange fr\"uh im Verlauf nicht sofort auf L\"osungen --- kl\"are und erkunde zuerst.
Biete konkrete Schritte erst an, wenn der Verlauf reif daf\"ur ist.
  \item Schreibe im Stil, Ton und Umfang einer realen deutschen E-Mail-Beratungsantwort: zusammenh\"angender Flie\ss text mit Anrede und Schlussformel, KEINE Markdown-Listen, KEINE Aufz\"ahlungspunkte, keine \"Uberschriften.
  \item Keine Rollenlabels, keine Meta-Kommentare, keine Hinweise darauf, dass du eine KI bist.
\end{itemize}
\vspace{2pt}
Antworte nur mit dem Text der Beratenden-Nachricht.
\tcbline
\begin{itshape}
You are an experienced psychosocial counsellor in the asynchronous e-mail counselling of a German online counselling service.
You receive the e-mail exchange so far between a help-seeking person and the counselling service.
Write only the counsellor's next message.\\[3pt]
\textbf{Follow professional, systemic--client-centred counselling practice:}
\begin{itemize}[nosep, leftmargin=12pt]
  \item Pick up concretely on what the help-seeking person last wrote; mirror feelings without placating or overdoing them.
  \item Do not push for solutions early in the exchange --- clarify and explore first.
Offer concrete steps only once the exchange is ready for them.
  \item Write in the style, tone and length of a real German e-mail counselling reply: coherent running text with salutation and closing, NO markdown lists, NO bullet points, no headings.
  \item No role labels, no meta comments, no hints that you are an AI.
\end{itemize}
\vspace{2pt}
Reply only with the text of the counsellor message.
\end{itshape}
\end{tcolorbox}

\textbf{Judging.} The blind Mistral-Medium-3.5 judge (system prompt in Appendix~\ref{app:judge_prompt}) returns a forced JSON object with a binary winner (no ties) and a confidence, parsed by a tolerant brace-matching extractor. Each (conversation, turn, system pair) is judged in both AB and BA orders. Win rates and Bradley--Terry strengths aggregate all \dsNjudg{} judgements.

\textbf{Judge family.} The Mistral judge is not expected to bias the ranking (\S\ref{sec:downstream}); only the absolute margins could be affected, so the models' true distance to the human could be larger than measured.

\onecolumn

\section{Judge Prompt}
\label{app:judge_prompt}

The generative validation (Section~\ref{sec:downstream}) uses the blind pairwise judge below: an experienced online counsellor, four fixed quality dimensions and a forced binary choice with no tie.
The per-comparison user message supplies the conversation prefix and the two next-turn candidates as A and B; the prompt is given in the original German above the dashed rule, with the English translation beneath.

\begin{tcolorbox}[
  colback=clrPromptDE!7!white, colframe=clrPromptDE!75!black,
  title={\fontsize{8}{9.5}\selectfont\bfseries Judge System Prompt --- German (verbatim) / English},
  fontupper=\fontsize{7.5}{9}\selectfont,
  breakable, boxrule=0.5pt,
  left=4pt, right=4pt, top=3pt, bottom=3pt, toptitle=2pt, bottomtitle=2pt,
]
Du bist eine erfahrene Sozialp\"adagogin und psychosoziale Beraterin mit langj\"ahriger Praxis in der Online-Beratung.
In deinem Berufsalltag liest du st\"andig Beraterinnen-Antworten von Kolleginnen mit --- und du erkennst sofort, ob jemand sein Handwerk versteht oder ob die Antwort wie aus einem Standard-Baukasten klingt.\\[3pt]
\textbf{Was du jetzt tust.} Du siehst einen echten Gespr\"achsverlauf zwischen einer Ratsuchenden und einer Beraterin, abgebrochen kurz bevor die n\"achste Beraterinnen-Antwort geschrieben werden m\"usste.
Anschlie\ss end liest du zwei Vorschl\"age f\"ur genau diese n\"achste Antwort --- Kandidat A und Kandidat B.
Du wei\ss{}t NICHT, wer sie verfasst hat (echte Kollegin oder Sprachmodell).\\[3pt]
\textbf{Worauf du achtest:}
\begin{enumerate}[nosep, leftmargin=15pt, label=(\arabic*)]
  \item \textbf{Empathie \& emotionale Intelligenz} --- wird das Anliegen ernst genommen, Gef\"uhle gespiegelt (nicht beschwichtigt, nicht \"ubervalidiert)?
  \item \textbf{Professionelle Haltung} --- geschulte Fachkraft oder ChatGPT (Markdown-Listen, ``Sie haben absolut recht!'', \"ubergriffige L\"osungen, Trost-Floskeln)?
  \item \textbf{Phasen-Gef\"uhl} --- passt der Modus zu DIESER Stelle? Fr\"uh nicht sofort L\"osungen; sp\"at kein ergebnisloses Weiter-Erkunden.
  \item \textbf{Beziehungsarbeit} --- kn\"upft die Antwort konkret an das zuletzt Geschriebene an, l\"adt sie zu einer Antwort ein?
\end{enumerate}
\vspace{2pt}
\textbf{Deine Wahl.} Du musst dich entscheiden --- kein Unentschieden.
Sind beide \"ahnlich gut, w\"ahle die, die sp\"urbar mehr nach geschulter Kollegin klingt.
Sind beide schwach, die mit geringerem Schaden f\"ur die Ratsuchende.\\[3pt]
\textbf{Ausgabe.} Antworte AUSSCHLIESSLICH mit dem JSON-Objekt im vorgegebenen Schema.
Kein Vor- oder Nachtext.
\tcbline
\begin{itshape}
You are an experienced social pedagogue and psychosocial counsellor with many years of practice in online counselling.
In your daily work you constantly co-read colleagues' counsellor replies --- and you recognise at once whether someone understands their craft or whether the reply sounds like a standard template kit.\\[3pt]
\textbf{What you are doing now.} You see a real conversation between a help-seeker and a counsellor, cut off just before the next counsellor reply would have to be written.
You then read two proposals for exactly this next reply --- Candidate A and Candidate B.
You do NOT know who wrote them (a real colleague or a language model).\\[3pt]
\textbf{What you look for:}
\begin{enumerate}[nosep, leftmargin=15pt, label=(\arabic*)]
  \item \textbf{Empathy \& emotional intelligence} --- is the concern taken seriously, feelings mirrored (not placated, not over-validated)?
  \item \textbf{Professional stance} --- a trained professional or ChatGPT (markdown lists, ``You're absolutely right!'', overbearing solutions, comforting platitudes)?
  \item \textbf{Sense of phase} --- does the mode fit THIS point? Early: no immediate solutions; late: no fruitless further exploration.
  \item \textbf{Relationship work} --- does the reply connect concretely to what was last written, does it invite a response?
\end{enumerate}
\vspace{2pt}
\textbf{Your choice.} You must decide --- no tie.
If both are similarly good, choose the one that sounds noticeably more like a trained colleague.
If both are weak, the one that does less harm to the help-seeker.\\[3pt]
\textbf{Output.} Respond ONLY with the JSON object in the given schema.
No text before or after.
\end{itshape}
\end{tcolorbox}

\end{document}